\documentclass[a4paper,11pt]{article}
\usepackage[T1]{fontenc}
\usepackage{lmodern}
\usepackage[margin=26mm]{geometry}
\usepackage{amsmath,amssymb,amsthm,mathtools}
\usepackage{microtype}
\usepackage{booktabs}
\usepackage{enumitem}
\usepackage{needspace}
\usepackage{xcolor}
\usepackage{hyperref}
\hypersetup{colorlinks=true,linkcolor=blue!45!black,citecolor=blue!45!black,
  urlcolor=blue!45!black,pdftitle={Singular wells and a unique Morse gate}}
\numberwithin{equation}{section}
\newtheorem{theorem}{Theorem}[section]
\newtheorem{proposition}[theorem]{Proposition}
\newtheorem{lemma}[theorem]{Lemma}
\newtheorem{corollary}[theorem]{Corollary}
\theoremstyle{definition}
\newtheorem{definition}[theorem]{Definition}
\newtheorem{assumption}[theorem]{Assumption}
\theoremstyle{remark}
\newtheorem{remark}[theorem]{Remark}
\newcommand{\R}{\mathbb R}
\newcommand{\T}{\mathbb T}
\newcommand{\E}{\mathbb E}
\newcommand{\eps}{\varepsilon}
\newcommand{\dd}{\,\mathrm d}
\newcommand{\cE}{\mathcal E}
\newcommand{\cL}{\mathcal L}
\DeclareMathOperator{\capc}{cap}
\DeclareMathOperator{\Vol}{Vol}
\DeclareMathOperator{\dist}{dist}
\DeclareMathOperator{\osc}{osc}
\DeclareMathOperator{\Int}{int}

\title{An Eyring-Kramers Formula for Singular Wells}
\author{
Yingli Wang\thanks{School of Mathematical Sciences, Fudan University, Shanghai, People's Republic of China; \texttt{yingliwang@fudan.edu.cn}}
\and
Lingjiong Zhu\thanks{Department of Mathematics, Florida State University, Tallahassee, Florida, United States of America; \texttt{zhu@math.fsu.edu}}}
\date{\today}

\begin{document}
\maketitle

\begin{abstract}
Understanding the persistence of learned behaviors in AI safety
motivates an investigation of the dynamical counterpart to the equilibrium geometry studied
in singular learning theory. We analyze this persistence through
transitions between singular wells.
We establish an Eyring-Kramers formula for a low-temperature Langevin
diffusion between two compact connected zero components of a
real-analytic potential.
The wells may be singular and nonisolated,
and the unique minimizing gate is a nondegenerate index-one saddle,
with gates understood as inclusion-minimal sets meeting every
minimizing path.
A boundary-gate construction shows that uniqueness of this
inclusion-minimal gate is equivalent to every minimizing path
passing through the saddle and one such path
meeting the communication level only there. Compact semianalytic topology
provides a separating function away from a fixed saddle neighborhood;
analytic Morse coordinates give matching capacity bounds by first
sending the temperature to zero and then shrinking that neighborhood.
An anchored energy estimate and a small-ball potential-theoretic
identity give uniformity for deterministic starting points. The leading
constant is expressed in terms of the sublevel-volume coefficient,
a Gamma factor, and the saddle Hessian. 
\end{abstract}

\section{Introduction}\label{sec:scope}

A central problem in AI safety is understanding when performance on
the training distribution translates into intended behavior in new
environments. Reward misspecification and distributional shift are
two sources of failure identified by Amodei et al.\ \cite{Amodei2016}.
Even when a training objective is appropriate, it can admit several
solutions with different behavior in deployment. Empirical work on
underspecification demonstrates that predictors with comparable
validation performance can differ substantially on deployment-relevant
tests \cite{DAmour2022}. In goal misgeneralization, a learned system
retains competence while pursuing an unintended goal in new situations
\cite{Shah2022}. These phenomena motivate understanding which solutions
training selects and how stable those solutions are under continued
optimization.

The stability question has a temporal aspect. If different regions
of a loss landscape correspond to different learned behaviors, noise
can induce transitions between them even after training has reached
a region of small loss. Predicting persistence then requires the
time scale of these transitions. \textit{Singular learning theory} (SLT) describes
how the geometry of a set of optimal parameters determines its
equilibrium weight \cite{Watanabe2009}, and its connection to internal
model structure and alignment motivates a corresponding theory of
learning dynamics \cite{PepinLehalleur2025}. 
\cite{MAISO74,MAISA7} isolate one precise part of this program:
determine the mean transition time from a singular well through
an ordinary Morse saddle under Langevin dynamics.
The equilibrium mass of a well describes how much probability it
receives at stationarity. Its mean transition time describes how long
optimization started there takes to reach another well. Our result
connects these quantities: the learning coefficient and its
multiplicity determine a power of temperature and a logarithmic
correction to the transition time, even when the barrier height is
fixed. For a learning problem in which separate behavioral analysis
identifies the wells with distinct learned solutions, this gives a
time scale for the persistence of those solutions under continued
noisy optimization. Establishing that correspondence involves the
model's predictions or internal representations in addition to its
training loss.

Delayed changes after fitting the training data provide one motivation
for studying such time scales. In grokking, generalization can improve
long after the training data have been fitted \cite{Power2022}.
Applying a transition-time formula to this phenomenon requires
identifying regions with different test performance and verifying
the dynamics and passage geometry between them. The mechanism
quantified here is a noise-induced transition between disconnected
optimal components separated by a positive loss barrier. It predicts
how singular well geometry changes the mean duration of residence
before such a transition. Motion within a connected set of minimizers
is another mechanism for changes after fitting; limiting SGD dynamics
near smooth manifolds of minimizers are studied in
\cite{LiWangArora2022}.

At low temperature, this is a metastability problem. Small random
perturbations allow a gradient system to move between regions that
its deterministic trajectories cannot connect. The Eyring--Kramers
law separates the exponential cost of crossing an energy barrier
from a prefactor determined by the geometry of the wells and the
passage between them. For isolated nondegenerate minima, this
prefactor is determined by Hessians. The geometry changes
substantially when a minimum is a connected set, or when the
Hessian loses rank within that set. We determine the resulting
temperature dependence while retaining a nondegenerate saddle
at the passage between two equally optimal components.

Many machine learning problems can be formulated as \emph{empirical risk minimization} (ERM): 
\begin{align}
\min_{x \in \mathbb{R}^d}  L(x),\qquad L(x):=\frac{1}{n}\sum_{i=1}^n \ell^{(i)}(x) \,, \label{eqn:optim}
\end{align}
where $L:\mathbb{R}^d\to\mathbb{R}$ denotes the loss function, $\ell^{(i)}$ the loss contributed by an individual data point $i \in \{1, \dots, n\}$, and $x\in\mathbb{R}^d$ the collection of all the parameters of the neural network.
In regression and classification problems such as logistic regression and support vector machines, $L$ is convex; whereas in deep learning $L$ is typically non-convex \cite{vapnik2013nature}. 
Since the number of data points is typically very large in modern machine learning applications, deterministic \textit{gradient descent} (GD) methods can be slow
\cite{bottou2010large}. 
\textit{Stochastic gradient descent} (SGD) is one of the most popular approaches for solving
optimization tasks such as \eqref{eqn:optim} with cheaper iteration costs based on the following iterates: 
\begin{equation}\label{eqn:sgd_main}
x_{k+1}=x_{k}-\eta\nabla\tilde{L}_{k}(x_{k}),
\qquad
\nabla\tilde{L}_k(x) := \frac{1}{b}\sum_{i\in \Omega_k} \ell^{(i)}(x),
\end{equation}
where $k$ denotes the iteration number, $\eta>0$ is the stepsize (learning rate), $\nabla\tilde{L}_{k}$ is the stochastic gradient at iteration $k$, 
and $\Omega_k \subset \{1,\dots,n\}$ denotes a random subset drawn from the set of data points with $|\Omega_k| = b \ll n$ for all $k$. 
SGD is widely used in \textit{deep learning} with a great success \cite{lecun2015deep,bottou2010large,bottou2008tradeoffs,DKLH2018,jastrzkebski2017three, zhang2015deep}.
A popular approach to understand the dynamics of stochastic optimization algorithms in deep learning is to impose some structure on the gradient noise $\nabla \tilde{L}_{k} (x_k) - \nabla L(x_k)$, 
and relate the process \eqref{eqn:sgd_main} to a stocahstic differential equation \cite{mandt2016variational,jastrzkebski2017three,hu2017diffusion,chaudhari2018stochastic,zhu2019anisotropic}. One might argue that if the batch-size $b$ is sufficiently large, 
then the gradient noise $\nabla \tilde{L}_{k} (x_k) - \nabla L(x_k)$ can be approximated by a Gaussian random vector ${\cal N}(0, \sigma^2 I_{d})$
due to the \textit{central limit theorem} (CLT) \cite{fischer2011history}. 
Gaussian approximations of the gradient noise has been proposed (see e.g. \cite{mandt2016variational}) so that \eqref{eqn:sgd_main} becomes:
\begin{equation}\label{eqn:sgd_gauss}
x_{k+1} = x_{k} - \eta \nabla L(x_k) + \sqrt{\eta} \sqrt{\eta \sigma^2} \xi_{k+1}, 
\end{equation}
where $\xi_k\sim\mathcal{N}(0,I_{d})$ are i.i.d. standard normal random variables in $\mathbb{R}^d$. If we further assume the stepsize $\eta$ is sufficiently small, the continuous-time analogue of \eqref{eqn:sgd_gauss} is the following \textit{overdamped Langevin diffusion}:
\begin{equation}\label{SDE:overdamped:Gaussian}
\mathrm{d}X_t = - \nabla L(X_t) \mathrm{d}t + \sqrt{2\varepsilon} \mathrm{d}B_t,
\end{equation}
where $\varepsilon:=\frac{\eta}{2}\sigma^2>0$ is the temperature parameter 
and $B_t$ is a standard Brownian motion in $\mathbb{R}^{d}$.
We note that more sophisticated approximations on gradient noise
have been made in terms of the covariance matrix of the Gaussian distribution (e.g.\ state dependent, anisotropic)
that results in more complicated SDE approximation than \eqref{SDE:overdamped:Gaussian} \cite{pmlr-v70-li17f,LiTaiE2019,Wang2017,Feng2019}; 
In paticular, Li, Tai, and E \cite{LiTaiE2019} establish weak approximations of SGD by
stochastic differential equations. The additive isotropic noise in
\eqref{SDE:overdamped:Gaussian} specifies a reversible model with the explicit Gibbs
measure, which makes a sharp mass-to-capacity
calculation possible.

We assume that $L\colon\T^d\to[0,\infty)$ is real-analytic, where $d\ge1$ and
$\T^d=(\R/2\pi\mathbb Z)^d$ has the flat metric, which is the setting
that is considered in \cite{MAISO74,MAISA7}.
The period $2\pi$ is a normalization: Remark~\ref{rem:period}
extends the result to $\T_m^d=(\R/2m\pi\mathbb Z)^d$ for every
fixed real $m>0$. Compactness makes the Gibbs normalization finite and removes questions of
escape to infinity, allowing us to isolate the effects of singular
wells and a Morse gate. Periodic parameter spaces also occur in
learning models: the quantum neural networks of
Girardi and De Palma \cite[Remark 2.2 and Section 3]{GirardiDePalma2025}
have a toroidal parameter space. Compact constraints also occur in
dictionary learning for image processing: Mairal et al.\
\cite[Section 2.1]{MairalBachPonceSapiro2010} constrain each dictionary
column to the closed Euclidean unit ball, so the dictionary variable
ranges over a compact set. Compact semianalytic parameter domains are
also part of the formulation of singular learning theory in
\cite[Section 3, Fundamental Conditions]{Watanabe2013}.
These examples motivate compact parameter domains; the diffusion
studied here specifically uses periodic parameters with a flat metric.
An extension to $\R^d$ requires a confining potential and estimates
on return times, Gibbs tails, and capacity outside large compact sets.
For an expanding family of tori, these estimates must be uniform in
$m$ to justify an $m\to\infty$ limit; the fixed-period rescaling in
Remark~\ref{rem:period} addresses each $m$ separately.
Assume that the zero
set has exactly two nonempty compact connected components $K_A$ and
$K_B$, so that
\begin{equation}\label{eq:zeros}
 L^{-1}(0)=K_A\cup K_B,\qquad K_A\cap K_B=\varnothing.
\end{equation}
In singular learning theory, the parameters representing an optimal
distribution can form a nonisolated analytic set with singularities;
neural networks, mixture models, and reduced-rank regression provide
examples of singular statistical models
\cite{Watanabe2009}; see also \cite[Sections 1 and 3]{Watanabe2013}.
Assumption~\eqref{eq:zeros} selects the case of two disconnected
optimal components. It separates the local singular geometry, which
determines the Gibbs mass near each component, from the positive
barrier that a continuous path between the components must cross.
This gives a basic setting in which to connect the equilibrium
quantities of singular learning theory with a transition time.

We consider the \textit{overdamped Langevin diffusion} on $\T^d$
at temperature $\eps>0$,
\begin{equation}\label{eq:sde}
 \mathrm dX_t=-\nabla L(X_t)\,\mathrm dt+\sqrt{2\eps}\,\mathrm dB_t,
 \qquad \cL_\eps=\eps\Delta-\nabla L\cdot\nabla,
\end{equation}
where $B_t$ is standard Brownian motion on the flat torus $\T^d$
and \eqref{eq:sde} is understood in local Euclidean coordinates.
The operator $\cL_\eps$ is the infinitesimal generator of $(X_t)_{t\geq0}$. Its
invariant probability measure is
\begin{equation}\label{eq:gibbs}
 \pi_\eps(\mathrm dw)=Z_\eps^{-1}e^{-L(w)/\eps}\dd w,
 \qquad Z_\eps=\int_{\T^d}e^{-L(w)/\eps}\dd w.
\end{equation}

A behavioral interpretation of the components in \eqref{eq:zeros}
additionally requires a comparison of their predictions or
representations. Distinct parameter components may represent the
same statistical model because of nonidentifiability.

Let $h$ be the infimum of $\max_{t\in[0,1]}L(\gamma(t))$ over continuous
paths joining $K_A$ to $K_B$. This value is positive: choose disjoint
open neighborhoods of the two zero components; their compact
complement has a positive minimum of $L$ and meets every connecting
path. Write $U$ and $U_B$ for the components of
$\{L<h\}$ containing $K_A$ and $K_B$, respectively. They are distinct:
otherwise a path in the open connected set $U$ would join the two zero
components with maximum strictly below $h$.

The sets $K_A$ and $K_B$ may have positive dimension and need not be
smooth submanifolds. Such geometries are relevant to singular
statistical models, where different parameters can represent the
same distribution and the loss can vanish on a singular set
\cite{Watanabe2009,Watanabe2013}. In this setting, a Gaussian
approximation at a single minimizer does not describe the mass of
an entire well. The appropriate local quantity is the volume of its
small sublevel sets. For a fixed neighborhood $U_A$ of $K_A$ whose
closure meets the zero set only in $K_A$, assume the sublevel asymptotic
\begin{equation}\label{eq:volume}
 V_A(s):=\Vol\{w\in U_A:L(w)<s\}
 \sim a_A s^{\lambda_A}(\log(1/s))^{m_A-1},
 \qquad s\downarrow0.
\end{equation}
Here $a_A>0$, $\lambda_A>0$, and $m_A\ge1$. The exponent
$\lambda_A$ and multiplicity $m_A$ describe the dominant singularity
of the whole component. In singular learning theory, $\lambda_A$
is the learning coefficient, or real log canonical threshold
\cite{Watanabe2009}. Lemma~\ref{lem:mass} identifies the corresponding
Gibbs-mass asymptotic; its leading coefficient differs from $a_A$
by the factor $\Gamma(\lambda_A+1)$.

We study the case in which the unique minimizing gate is a single
nondegenerate index-one critical point $z$. 
The precise
potential-theoretic definition is given in Definition~\ref{def:gate},
and the single-gate condition is stated in Assumption~\ref{ass:single-gate}.
Let
\begin{equation}\label{eq:cz}
 L(z):=h,\quad H_z:=\nabla^2L(z),\quad
 -\mu_-<0\text{ be its negative eigenvalue},\quad
 c_z:=\frac{\mu_-(2\pi)^{d/2-1}}{\sqrt{|\det H_z|}}.
\end{equation}
For a small fixed $\rho>0$, let $B_\rho$ be the closed
$\rho$-neighborhood of $K_B$ and let $\tau_{B_\rho}$ be its hitting
time. The question posed in 
\cite{MAISO74,MAISA7} is to determine its mean from the well-volume
law and the saddle geometry. Theorem~\ref{thm:main} gives the sharp
asymptotic, uniformly for $x$ in each compact subset of $U$. The power
$\lambda_A-d/2$ compares the mass near the initial well with the
capacity of the saddle, and the factor
$(\log(1/\eps))^{m_A-1}$ records the multiplicity of the well
singularity. When $K_A=\{x_A\}$ is a nondegenerate minimum, Gaussian
integration gives $\lambda_A=d/2$, $m_A=1$, and
$a_A\Gamma(d/2+1)=(2\pi)^{d/2}/\sqrt{\det\nabla^2L(x_A)}$.
Consequently, Theorem~\ref{thm:main} recovers the classical formula
\cite[equation (3.3)]{BEGK2004}
\begin{equation}\label{eq:classical-ek}
 \E_x\left[\tau_{B_\rho}\right]\sim
 \frac{2\pi}{\mu_-}
 \sqrt{\frac{|\det H_z|}{\det\nabla^2L(x_A)}}\,e^{h/\eps}.
\end{equation}
For a singular well, the logarithm of the asymptotic in
Theorem~\ref{thm:main} is
\begin{equation}\label{eq:log-transition}
 \log\E_x\left[\tau_{B_\rho}\right]
 =\frac h\eps+\left(\lambda_A-\frac d2\right)\log\eps
  +(m_A-1)\log\log(1/\eps)+\log C_A+o(1).
\end{equation}
With the dimension and barrier fixed, a smaller $\lambda_A$ gives
a longer mean transition time at sufficiently low temperature;
at equal $\lambda_A$, a larger $m_A$ gives a further logarithmic
increase. Equation~\eqref{eq:log-transition} makes the dynamical
contribution of the equilibrium volume data explicit.
The result gives a mean first-passage time under
\eqref{eq:sde}; estimating the probability of a transition within
a prescribed training horizon requires additional information
about the hitting-time distribution.

\subsection{Related work}

The AI safety motivation combines work on generalization failures with
work on the selection of learned solutions. D'Amour et al.\
\cite{DAmour2022} document the ambiguity left by training and validation
criteria across several application domains. Shah et al.\
\cite{Shah2022} demonstrate that an unintended goal can generalize
beyond the situations in which it produced good training performance.
Pepin Lehalleur et al.\ \cite{PepinLehalleur2025} argue that connecting
data distributions, internal model structure, and generalization is
central to alignment, and discuss singular learning theory as a
framework for internal model selection. The transition-time question
studied here adds a dynamical quantity to this geometric perspective:
the time required to move between specified components of the
optimal set.

For stochastic gradient descent (SGD), the covariance of the gradient
noise also affects selection. Li, Wang, and Arora
\cite{LiWangArora2022} derive limiting dynamics near a smooth manifold
of minimizers, with coefficients determined jointly by the loss
geometry and the noise covariance. Their framework describes motion
along the manifold after the iterates approach it. Here the process
has additive isotropic noise, and the quantity of interest is the
rare transition between distinct components. Extending the present
formula to SGD requires control of its state-dependent noise and of
the diffusion approximation on the relevant transition time scale.

The physical origins of the transition-time law lie in Eyring's
transition-state theory \cite{Eyring1935} and Kramers' diffusion
model of barrier crossing \cite{Kramers1940}. Freidlin--Wentzell
theory \cite{FreidlinWentzell2012} places small-noise exit problems
in a large-deviation framework and identifies their exponential
time scales. Determining a relative-error asymptotic such as
\eqref{eq:classical-ek} also requires the subexponential factor.
Bovier, Eckhoff, Gayrard, and Klein \cite{BEGK2004} developed a
potential-theoretic approach that relates mean exit times to
capacities and computes the latter from the local geometry of
nondegenerate saddles. Bovier, Gayrard, and Klein \cite{BGK2005}
subsequently related the low-lying spectrum to a capacity matrix
and to inverse mean metastable exit times under generic assumptions
on the well depths. Our argument uses the same variational and
potential-theoretic framework, with estimates adapted to connected
singular sets of minima.

Degenerate critical points have been treated through several
extensions of this framework. Berglund and Gentz
\cite{BerglundGentz2010} derived sharp prefactors for nonquadratic
saddles and studied their behavior near bifurcations. Their results
show explicitly how vanishing curvatures change the temperature
dependence of the transition law. Avelin, Julin, and Viitasaari
\cite{AvelinJulinViitasaari2023} obtained a geometric characterization
of capacity using weighted distance and separating-surface
quantities, together with transition-time upper bounds allowing
degenerate minima and saddles and several saddles at the same
height. These works emphasize the role of local passage geometry
and the global arrangement of the transition routes.

Spectral methods also extend beyond isolated Morse critical points.
Assal, Bony, and Michel \cite{AssalBonyMichel2025} established
Eyring--Kramers asymptotics for the small eigenvalues of the Witten
Laplacian associated with confining Morse--Bott potentials. In that
setting, critical sets are smooth submanifolds with nondegenerate
normal Hessians. Delande \cite{Delande2026} treated classes of
isolated degenerate critical points admitting diagonal monomial
normal forms and obtained sharp spectral asymptotics with explicit
powers of the semiclassical parameter. The well sets considered
here may be singular, and the Hessian of $L$ may change rank along them.
The example in Appendix~\ref{sec:model} has a circle of minima whose
transverse quadratic coefficient vanishes at one point; this
produces a logarithmic factor in the mean transition time.

The volume asymptotic \eqref{eq:volume} belongs to the analytic
framework of singular learning theory developed by Watanabe
\cite{Watanabe2009,Watanabe2013}. Resolution of singularities
explains the powers and logarithms in integrals concentrated near
analytic zero sets and replaces the dimension-based Gaussian
scaling by learning coefficients and their multiplicities.
Here these equilibrium quantities enter a dynamical formula
through the mass-to-capacity ratio. The remaining steps are to
localize capacity at the unique gate despite the singular wells,
and to pass from an equilibrium boundary average to a
deterministic initial point. For the first step, the semianalytic
topology developed by Bierstone and Milman
\cite{BierstoneMilman1988} provides the finiteness and accessibility
properties used in the global separation argument. For the second,
we combine an anchored energy estimate with local elliptic
estimates. This proves the singular-well formula requested in
\cite{MAISO74,MAISA7} under the gate condition stated below.

\paragraph{Our Contribution}
The main contributions are as follows.
\begin{enumerate}[label=(\roman*),leftmargin=2em]
 \item \textit{A sharp transition law for analytic singular wells.}
 Theorem~\ref{thm:main} gives the transition-time asymptotic, including
 the explicit constant and uniformity over compact subsets of $U$.
 The well geometry enters through $(a_A,\lambda_A,m_A)$, and the
 saddle geometry enters through $(\mu_-,H_z)$.
 \item \textit{Capacity localization under a unique-gate condition.}
 Theorem~\ref{thm:general-capacity} gives the Morse-saddle capacity
 for arbitrary compact connected analytic zero components.
 Lemmas~\ref{lem:boundary-gate} and~\ref{lem:gate-criterion}
 identify the global connections forced by gate uniqueness, and
 Corollary~\ref{cor:critical-level} provides a sufficient condition
 stated directly in terms of critical points at height $h$.
 \item \textit{Uniform control for deterministic initial states.}
 Theorem~\ref{thm:reduction} converts one fixed-neighborhood
 capacity asymptotic into a transition law uniform on every
 compact $Q\subset U$. Corollary~\ref{cor:target} proves that the
 leading constant is unchanged for any fixed closed target
 neighborhood compactly contained in $U_B$.
 \item \textit{An explicit logarithmic example.}
 Propositions~\ref{prop:model-mass} and~\ref{prop:gate} compute the
 well volume and the unique gate for a two-dimensional analytic
 potential with a non-Morse--Bott well.
 Corollary~\ref{cor:example} yields the
 prefactor $\eps^{-1/2}\log(1/\eps)$ and its explicit coefficient.
\end{enumerate}

\paragraph{Organization of the paper}
Section~\ref{sec:results} contains the main results and their proofs.
In particular,
Section~\ref{sec:gate-main} states the gate hypothesis and the
transition-time theorem. Sections~\ref{sec:mass} and~\ref{sec:reduction}
derive the Gibbs-mass asymptotic and the reduction to deterministic
initial states. Section~\ref{sec:general} proves the gate criteria
and the capacity asymptotic, completing the proof of the main
theorem. Section~\ref{sec:conclusion}
concludes and discusses further directions.
Appendix~\ref{sec:model} studies an explicit logarithmic
example.

\subsection{Notation}\label{sec:notation}
All gradients, Hessians, distances, and volumes on $\T^d$ refer to
the flat metric, and $\dd w$ denotes its volume element. We write
$\Int D$ for the interior of $D$.
Let $\mathfrak p\colon\R^d\to\T^d$ be the quotient map.
For $x,y\in\T^d$ with representatives $\widetilde x,\widetilde y\in\R^d$,
and nonempty subsets $E,F\subset\T^d$, define
\begin{equation}\label{eq:flat-distance}
 \begin{split}
 \dist(x,y)&:=\min_{k\in\mathbb Z^d}
                 |\widetilde x-\widetilde y+2\pi k|,\\
 \dist(x,E)&:=\inf_{y\in E}\dist(x,y),\qquad
 \dist(E,F):=\inf_{x\in E,\,y\in F}\dist(x,y).
 \end{split}
\end{equation}
Here $|\cdot|$ is the Euclidean norm. A flat coordinate chart centered
at $x$ is the map $\xi\mapsto\mathfrak p(\widetilde x+O\xi)$
on a Euclidean ball of radius less than $\pi$, where $O$ is an
orthogonal matrix. Its metric tensor is the identity, so gradients,
Hessians, and volume elements in this chart are Euclidean.
We use $x+\xi$ for this chart when $O=I$ and write $B_r(x)$ for
the corresponding metric ball.
The gradient-flow basin is
$\mathcal B_A=\{x:\dist(\phi_t(x),K_A)\to0\text{ as }t\to\infty\}$,
where $\dot\phi_t=-\nabla L(\phi_t)$. For a closed set $B$,
$\tau_B:=\inf\{t\ge0:X_t\in B\}$, and $\E_x$ denotes expectation
with $X_0=x$. We write $f_\eps\sim g_\eps$ if
$f_\eps/g_\eps\to1$ as $\eps\downarrow0$; uniformity is stated
explicitly when additional parameters vary. Constants denoted by
$C$ may change from line to line and are independent of $\eps$
unless stated otherwise. We use
$\Gamma(a):=\int_0^\infty t^{a-1}e^{-t}\dd t$ for $a>0$.
For an open set $D\subset\T^d$, $H^1(D)=W^{1,2}(D)$ is the space
of $L^2(D)$ functions whose distributional first derivatives in flat
charts belong to $L^2(D)$, with norm
\[
 \|f\|_{H^1(D)}^2:=\int_D\left(|f|^2+|\nabla f|^2\right)\dd w.
\]
On $\T^d$ this is equivalently the space of $2\pi$-periodic Sobolev
functions with finite norm on a fundamental cell.
We write $H_0^1(D)$ for the closure of $C_c^\infty(D)$ in $H^1(D)$;
the same definitions apply to Euclidean domains with Lebesgue measure.
For a real symmetric matrix $H\in\R^{n\times n}$, $n\ge1$,
$\lambda_{\min}(H)$ and $\lambda_{\max}(H)$ denote its smallest
and largest eigenvalues. Thus
\[
 \lambda_{\min}(H)|v|^2\le v^{\top}Hv
 \le\lambda_{\max}(H)|v|^2,\qquad v\in\R^n.
\]

\section{Main Results}\label{sec:results}

\subsection{Gate hypothesis and transition-time formula}\label{sec:gate-main}

We use the potential-theoretic meaning of the single-gate wording,
with a gate defined as in \cite[Definition 1.1]{BEGK2004}.
Minimality below is among all subsets, without a closedness restriction.

\begin{definition}[Level cuts and gates]\label{def:gate}
Let
\[
 \mathcal P:=\left\{\gamma\in C([0,1];\T^d):
       \gamma(0)\in K_A,\ \gamma(1)\in K_B\right\},\qquad
 \mathcal P_h:=\left\{\gamma\in\mathcal P:
       \max_{t\in[0,1]}L(\gamma(t))=h\right\}.
\]
A level cut is a set
$G\subset\{L=h\}$ satisfying
\[
 G\cap\gamma([0,1])\ne\varnothing
 \quad\text{for every }\gamma\in\mathcal P_h.
\]
A gate $G$ is an inclusion-minimal level cut, i.e. $G$ is a level cut and
every proper subset $G'\subsetneq G$ fails to be a level cut, meaning
that some $\gamma\in\mathcal P_h$ satisfies
$G'\cap\gamma([0,1])=\varnothing$. 
\end{definition}

In Definition~\ref{def:gate}, 
for each $\gamma\in\mathcal P$, continuity of $L\circ\gamma$ on
$[0,1]$ ensures that its maximum is attained. Thus
$\gamma\in\mathcal P_h$ means that $L(\gamma(t))\le h$ for every
$t\in[0,1]$ and $L(\gamma(t_*))=h$ for some $t_*\in[0,1]$.
Moreover, the family $\mathcal P_h$ in Definition~\ref{def:gate}
is nonempty by
Lemma~\ref{lem:tame-topology}.

The capacity calculation uses the following condition on the gates
between the two zero components.

\begin{assumption}[Single minimizing gate]\label{ass:single-gate}
There is exactly one gate in the sense of Definition~\ref{def:gate},
and it is $\{z\}$, where $z$ is a nondegenerate index-one critical
point of $L$ at height $h$, i.e.
$L(z)=h$, $\nabla L(z)=0$,
and $H_z=\nabla^2L(z)$ has one negative and $d-1$
positive eigenvalues.
\end{assumption}

Lemma~\ref{lem:boundary-gate} constructs gates on the boundaries of
the two strict sublevel components. Lemma~\ref{lem:gate-criterion}
then proves that this hypothesis is equivalent
to two path statements: every minimizing path meets $z$, and some
minimizing path has $z$ as its only point at height $h$. This specifies
uniqueness of the relevant gate and permits other, irrelevant critical
points, including degenerate ones at height $h$. It also proves that
using minimal closed level cuts would give the same single-gate
condition in this analytic setting.

The main theorem combines the well-volume data with this gate condition.
Its uniformity range includes every compact subset of the gradient-flow
basin below the communication height.

\begin{theorem}[Singular-well Eyring--Kramers formula]\label{thm:main}
Let $L\colon\T^d\to[0,\infty)$ be real-analytic and satisfy
\eqref{eq:zeros}, \eqref{eq:volume}, and
Assumption~\ref{ass:single-gate}. For sufficiently small fixed $\rho>0$, let
\[
 B_\rho:=\left\{w\in\T^d:\dist(w,K_B)\le\rho\right\},\qquad
 \tau_{B_\rho}:=\inf\left\{t\ge0:X_t\in B_\rho\right\}.
\]
For every nonempty compact $Q\subset U$,
\begin{equation}\label{eq:main-uniform}
 \lim_{\eps\downarrow0}\sup_{x\in Q}
 \left|\frac{\E_x[\tau_{B_\rho}]}
 {C_A\eps^{\lambda_A-d/2}
 (\log(1/\eps))^{m_A-1}e^{h/\eps}}-1\right|=0,
\end{equation}
where
\begin{equation}\label{eq:main-constant}
 C_A:=\frac{a_A\Gamma(\lambda_A+1)\sqrt{|\det H_z|}}
 {\mu_-(2\pi)^{d/2-1}}.
\end{equation}
In particular, \eqref{eq:main-uniform} holds for the compact subsets of
$\Int(\mathcal B_A\cap\{L<h\})$.
\end{theorem}

\begin{proof}
The proof of Theorem~\ref{thm:main} is completed in Section~\ref{sec:general}. 
\end{proof}

The rescaling discussion in the following remark 
describes how Theorem~\ref{thm:main} applies when the
period of the flat torus changes.

\begin{remark}[Any fixed positive period]\label{rem:period}
Let $m>0$ be fixed and let $L_m$ on $\T_m^d$ satisfy the hypotheses
of Theorem~\ref{thm:main}, with distances and volumes measured in its
flat metric. Define $F_m(x):=mx$ from $\T^d$ to $\T_m^d$ and
$\widetilde L:=L_m\circ F_m$. If $X_t$ is the Langevin diffusion
for $L_m$, then $Y_t:=F_m^{-1}(X_{m^2t})$ has generator
$\eps\Delta-\nabla\widetilde L\cdot\nabla$. Write $V_{A,m}$ for
the sublevel-volume function in \eqref{eq:volume} on $\T_m^d$
and $\widetilde V_A$ for the volume in the pulled-back neighborhood
on $\T^d$. For a closed target $T\subset\T_m^d$, Brownian scaling
and the change of variables $w=mx$ give
\begin{equation}\label{eq:period-scaling}
 \tau_T(X)=m^2\tau_{F_m^{-1}(T)}(Y),\qquad
 \widetilde V_A(s)=m^{-d}V_{A,m}(s),\qquad
 \nabla^2\widetilde L(F_m^{-1}(z))=m^2\nabla^2L_m(z).
\end{equation}
The map $F_m$ preserves the communication height and the gate
conditions. If $a_{A,m}$ is the volume coefficient for $L_m$,
$H_{z,m}:=\nabla^2L_m(z)$, and $-\mu_{-,m}$ is its negative
eigenvalue, then \eqref{eq:period-scaling} implies
\[
 \widetilde a_A=m^{-d}a_{A,m},\qquad
 \widetilde\mu_-=m^2\mu_{-,m},\qquad
 \sqrt{|\det\widetilde H_z|}=m^d\sqrt{|\det H_{z,m}|}.
\]
Let $\widetilde C_A$ be the constant \eqref{eq:main-constant}
for $\widetilde L$. Consequently $m^2\widetilde C_A=C_{A,m}$, where
\[
 C_{A,m}:=\frac{a_{A,m}\Gamma(\lambda_A+1)\sqrt{|\det H_{z,m}|}}
                   {\mu_{-,m}(2\pi)^{d/2-1}}.
\]
Theorem~\ref{thm:main} therefore has the same form on $\T_m^d$,
including uniformity over each fixed compact subset of its source
component. Here $m$ remains fixed as $\eps\downarrow0$.
\end{remark}

First we establish
the mass asymptotic and a reduction from fixed-set capacity to
deterministic starting points. The general capacity proof combines
analytic topology with the local geometry of the unique Morse gate.

\subsection{Sublevel volume and Gibbs mass}\label{sec:mass}

The conversion of \eqref{eq:volume} to Gibbs mass fixes the Gamma factor
in the eventual constant.

\begin{lemma}[Mass from volume]\label{lem:mass}
Under \eqref{eq:volume},
\begin{equation}\label{eq:mass}
 M_A(\eps):=\int_{U_A}e^{-L(w)/\eps}\dd w
 \sim c_A\eps^{\lambda_A}(\log(1/\eps))^{m_A-1},
 \qquad c_A:=a_A\Gamma(\lambda_A+1).
\end{equation}
Replacing $U_A$ by any other fixed neighborhood whose closure meets the
zero set exactly in $K_A$ preserves this asymptotic.
\end{lemma}

\begin{proof}
Tonelli's theorem applied to
$e^{-L(w)/\eps}=\eps^{-1}\int_{L(w)}^\infty e^{-s/\eps}\dd s$ gives
\begin{equation}\label{eq:layercake}
 M_A(\eps)=\frac1\eps\int_0^\infty e^{-s/\eps}V_A(s)\dd s
 =\int_0^\infty e^{-u}V_A(\eps u)\dd u.
\end{equation}
Let $r:=m_A-1$. By \eqref{eq:volume}, there is
$s_0\in(0,e^{-1})$ such that
\begin{equation}\label{eq:volume-two-sided}
 \frac{a_A}{2}s^{\lambda_A}(\log(1/s))^r
 \le V_A(s)\le
 2a_A s^{\lambda_A}(\log(1/s))^r,
 \qquad 0<s<s_0.
\end{equation}
The upper bound in \eqref{eq:volume-two-sided} gives, for
$0<u<s_0/\eps$ and $\eps<\min(s_0,e^{-1})$,
\[
 \frac{V_A(\eps u)}{\eps^{\lambda_A}(\log(1/\eps))^r}
 \le C u^{\lambda_A}(1+|\log u|)^r.
\]
The right-hand side above is integrable against $e^{-u}\dd u$ on $(0,\infty)$.
The remainder satisfies
\[
 \int_{s_0/\eps}^\infty e^{-u}V_A(\eps u)\dd u
 \le\Vol(U_A)e^{-s_0/\eps}.
\]
The dominated convergence theorem in \eqref{eq:layercake} proves
\eqref{eq:mass}.
For a second allowed neighborhood $\widetilde U_A$, compactness of
$K_A$ provides an open set $U_0$ such that
\[
 K_A\subset U_0,\qquad \overline U_0\subset U_A\cap\widetilde U_A.
\]
The set $F:=(\overline U_A\cup\overline{\widetilde U_A})\setminus U_0$
is compact. The conditions
\[
 \overline U_A\cap L^{-1}(0)
 =\overline{\widetilde U_A}\cap L^{-1}(0)=K_A,
 \qquad K_A\subset U_0,
\]
imply
\[
 F\cap L^{-1}(0)=\varnothing,\qquad
 \overline{U_A\triangle\widetilde U_A}\subset F.
\]
If $F$ is nonempty, let $c:=\min_{w\in F}L(w)>0$. It follows that
\[
 \left|\int_{U_A}e^{-L/\eps}\dd w
       -\int_{\widetilde U_A}e^{-L/\eps}\dd w\right|
 \le\int_{U_A\triangle\widetilde U_A}e^{-L/\eps}\dd w
 \le\Vol(\T^d)e^{-c/\eps}=o(M_A(\eps)).
\]
If $F$ is empty, the two integrals agree. This proves the
neighborhood independence in both cases.
\end{proof}

\subsection{A fixed-capacity reduction for singular wells}\label{sec:reduction}

All capacities in this paper use the unnormalized weight
$w_\eps:=e^{-L/\eps}$. A smooth closed neighborhood means the closure
of an open neighborhood with smooth boundary. For two disjoint such
sets $A,B$, define Dirichlet form and capacity as:
\begin{align}
 \cE_\eps(f,g)&:=\eps\int_{\T^d}\nabla f\cdot\nabla g\,w_\eps\dd w,
 \label{eq:energy}\\
 \capc_\eps(A,B)&:=\inf\left\{\cE_\eps(f,f):f\in H^1(\T^d),\
                 f|_A=1,\ f|_B=0\right\}.
 \label{eq:cap}
\end{align}
For a measurable region $D\subset\T^d$, we also write
\[
\cE_{\eps,D}(f,g):=\eps\int_D\nabla f\cdot\nabla g\,w_\eps\dd w.
\]
We record the Poincar\'e inequality needed for \eqref{eq:cap}.
For $v\in H^1(\T^d)$, let
\[
 \widehat v(k):=(2\pi)^{-d}\int_{\T^d}
     v(w)e^{-\mathrm{i}k\cdot w}\dd w,\qquad
 k\in\mathbb Z^d,\qquad \bar v:=\widehat v(0).
\]
Parseval's identity gives
\begin{equation}\label{eq:torus-poincare}
 \|v-\bar v\|_2^2
 =(2\pi)^d\sum_{k\in\mathbb Z^d\setminus\{0\}}|\widehat v(k)|^2
 \le(2\pi)^d\sum_{k\in\mathbb Z^d}|k|^2|\widehat v(k)|^2
 =\|\nabla v\|_2^2.
\end{equation}
If $v=0$ on $S:=A\cup B$, then
$|S|^{1/2}|\bar v|=\|v-\bar v\|_{L^2(S)}$.
The triangle inequality and \eqref{eq:torus-poincare} imply
\begin{equation}\label{eq:torus-anchored-poincare}
 \|v\|_2\le K_{A,B}\|\nabla v\|_2,\qquad
 K_{A,B}:=1+\left(\frac{\Vol(\T^d)}{|A\cup B|}\right)^{1/2}.
\end{equation}
For each fixed $\eps>0$, \eqref{eq:energy} and
\eqref{eq:torus-anchored-poincare} give
\begin{equation}\label{eq:capacity-coercivity}
 \|v\|_{H^1(\T^d)}^2
 \le \left(1+K_{A,B}^2\right)\eps^{-1}e^{\max_{\T^d}L/\eps}
       \cE_\eps(v,v),\qquad v|_{A\cup B}=0.
\end{equation}
The admissible affine space
\[
 \mathcal A_{A,B}:=\left\{f\in H^1(\T^d):f|_A=1,\ f|_B=0\right\}
\]
in \eqref{eq:cap} is nonempty and weakly closed in $H^1(\T^d)$.
The functional $f\mapsto\cE_\eps(f,f)$ is weakly lower
semicontinuous, since
\[
 \cE_\eps(f,f)=\left\|\sqrt{\eps w_\eps}\,\nabla f\right\|_{L^2(\T^d;\R^d)}^2
 \le\liminf_{n\to\infty}\cE_\eps(f_n,f_n)
 \quad\text{if }f_n\rightharpoonup f\text{ in }H^1(\T^d).
\]
It is strictly convex on $\mathcal A_{A,B}$: for distinct
$f,g\in\mathcal A_{A,B}$ and $0<\theta<1$,
\begin{align*}
 \cE_\eps(\theta f+(1-\theta)g,\theta f+(1-\theta)g)
 &=\theta\cE_\eps(f,f)+(1-\theta)\cE_\eps(g,g)
   -\theta(1-\theta)\cE_\eps(f-g,f-g)\\
 &<\theta\cE_\eps(f,f)+(1-\theta)\cE_\eps(g,g).
\end{align*}
Here \eqref{eq:capacity-coercivity} implies
$\cE_\eps(f-g,f-g)>0$, because $f-g=0$ on $A\cup B$.
Fixing $f_0\in\mathcal A_{A,B}$, the bound
$\cE_\eps(f-f_0,f-f_0)\le2\cE_\eps(f,f)+2\cE_\eps(f_0,f_0)$
and \eqref{eq:capacity-coercivity} make every minimizing sequence
bounded in $H^1(\T^d)$. Weak lower semicontinuity therefore gives
a minimizer of \eqref{eq:cap}, and strict convexity gives its
uniqueness. The coercivity constant here may depend on $\eps,A,B$.
The minimizer $q_{A,B}$ is extended as $1$ on $A$ and $0$ on $B$.
Truncation in \eqref{eq:cap} gives $0\le q_{A,B}\le1$, and the Euler
equation is $\cL_\eps q_{A,B}=0$ outside $A\cup B$.

The following theorem isolates the dynamical step. Its capacity
assumption concerns just one fixed pair of neighborhoods.

\begin{theorem}[Capacity implies the pointwise transition law]
\label{thm:reduction}
Assume \eqref{eq:zeros} and \eqref{eq:volume}. Let $A_*\subset U$ and
$B\subset U_B$ be closed neighborhoods of $K_A$ and $K_B$ with smooth
boundaries. Suppose that for some $c_*>0$,
\begin{equation}\label{eq:cap-assumption}
 \capc_\eps(A_*,B)=c_*\eps^{d/2}e^{-h/\eps}(1+o(1)).
\end{equation}
For every nonempty compact $Q\subset U$,
\begin{equation}\label{eq:reduction-conclusion}
 \lim_{\eps\downarrow0}\sup_{x\in Q}
 \left|\frac{\capc_\eps(A_*,B)\,\E_x[\tau_B]}{M_A(\eps)}-1\right|=0.
\end{equation}
Here $M_A$ is defined in \eqref{eq:mass}. The proof only needs $L\in C^3$
in addition to the volume asymptotic \eqref{eq:volume} and the zero-set
condition \eqref{eq:zeros}.
\end{theorem}

\begin{proof}
The proof of Theorem~\ref{thm:reduction} follows
Lemmas~\ref{lem:anchor} and~\ref{lem:flat} below in this subsection.
\end{proof}

The proof of Theorem~\ref{thm:reduction} uses two elementary estimates. The first turns small weighted
energy into near constancy on a fixed region below the barrier.

\begin{lemma}[Anchoring a capacitary potential]\label{lem:anchor}
Let $D$ be a bounded connected smooth domain with $\overline D\cap B
=\varnothing$, and let $D_0$ have compact closure contained in $D$.
Suppose $L\le b$ on $D$,
$q=q_{A,B}$, and $S\subset A\cap D$ is measurable with $|S|>0$.
Then, for all sufficiently small $\eps$,
\begin{align}
 \|1-q\|_{L^2(D)}
 &\le C_D\left(1+|S|^{-1/2}\right)
 \left(\eps^{-1}e^{b/\eps}\capc_\eps(A,B)\right)^{1/2},
 \label{eq:anchor-l2}\\
 \sup_{w\in D_0}(1-q(w))
 &\le C_{D,D_0}\eps^{-d/2}\left(1+|S|^{-1/2}\right)
 \left(\eps^{-1}e^{b/\eps}\capc_\eps(A,B)\right)^{1/2}.
 \label{eq:anchor-sup}
\end{align}
The constant $C_D$ depends only on $D$ and $d$;
$C_{D,D_0}$ may also depend on $D_0$ and $\|\nabla L\|_\infty$.
Both constants are independent of $A,S,\eps$.
If instead $\overline D\cap A=\varnothing$, $L\le b$ on $D$, and
$S_B\subset B\cap D$ is measurable with $|S_B|>0$, then
\begin{align}
 \|q\|_{L^2(D)}
 &\le C_D\left(1+|S_B|^{-1/2}\right)
 \left(\eps^{-1}e^{b/\eps}\capc_\eps(A,B)\right)^{1/2},
 \label{eq:zero-anchor-l2}\\
 \sup_{w\in D_0}q(w)
 &\le C_{D,D_0}\eps^{-d/2}\left(1+|S_B|^{-1/2}\right)
 \left(\eps^{-1}e^{b/\eps}\capc_\eps(A,B)\right)^{1/2}.
 \label{eq:zero-anchor-sup}
\end{align}
\end{lemma}

\begin{proof}
For a smooth connected domain $D\subset\T^d$, the mean-zero
Poincar\'e inequality is
\begin{equation}\label{eq:domain-poincare}
 \|f-f_D\|_{L^2(D)}\le P_D\|\nabla f\|_{L^2(D)},\qquad
 f_D:=|D|^{-1}\int_Df\dd w,\qquad f\in H^1(D).
\end{equation}
Its Euclidean form is \cite[Section 5.8.1, Theorem 1]{Evans2010}.
For completeness, the same inequality holds when $D$ crosses several
torus charts. A finite smooth coordinate cover of $\overline D$,
a subordinate partition of unity, and boundary flattening reduce
compactness of $H^1(D)\hookrightarrow L^2(D)$ to the extension
theorem and the Rellich--Kondrachov compactness theorem in
\cite[Section 5.4, Theorem 1; Section 5.7, Theorem 1]{Evans2010}.
The one-dimensional compact embedding also follows from
$|f(x)-f(y)|\le |x-y|^{1/2}\|f'\|_2$ on coordinate intervals.
If \eqref{eq:domain-poincare} failed, normalization would give
\[
 \int_D f_n\dd w=0,\qquad \|f_n\|_{L^2(D)}=1,\qquad
 \|\nabla f_n\|_{L^2(D)}\le n^{-1}.
\]
A subsequence converges strongly in $L^2(D)$ to $f$, with
$\nabla f=0$ in distributions, $\int_Df\dd w=0$, and
$\|f\|_{L^2(D)}=1$. Connectedness makes $f$ constant, contradicting
the last two identities. This proves \eqref{eq:domain-poincare}.

Let $v:=1-q$ and $v_D:=|D|^{-1}\int_Dv\dd w$. Since $v=0$ on $S$,
\eqref{eq:domain-poincare} and the triangle inequality give
\begin{align*}
 |S|^{1/2}|v_D|&\le\|v-v_D\|_{L^2(S)}
 \le P_D\|\nabla v\|_{L^2(D)},\\
 \|v\|_{L^2(D)}&\le\|v-v_D\|_{L^2(D)}+|D|^{1/2}|v_D|\le P_D\left(1+|D|^{1/2}|S|^{-1/2}\right)
       \|\nabla v\|_{L^2(D)},\\
 \int_D|\nabla v|^2\dd w
 &\le \eps^{-1}e^{b/\eps}\cE_\eps(q,q).
\end{align*}
These inequalities prove \eqref{eq:anchor-l2}.

The extension of $v$ by zero across $A$ is a nonnegative weak
subsolution of $\operatorname{div}(w_\eps\nabla v)\ge0$ in $D$.
Indeed, let $n_A(w)$ be the outward unit normal to $A$ at
$w\in\partial A\cap D$. For sufficiently small $t>0$, the point
$w+t n_A(w)$ lies in $D\setminus A$, and boundary regularity gives
\[
 \partial_{n_A}v(w)
 =\lim_{t\downarrow0}\frac{v(w+t n_A(w))-v(w)}{t}\ge0,
\]
since $v(w)=0$ and $v\ge0$ outside $A$.
The outward unit normal to $D\setminus A$ on $\partial A\cap D$
is $-n_A$. For $\psi\in C_c^\infty(D)$ with $\psi\ge0$,
integration by parts on $D\setminus A$ yields
\begin{equation}\label{eq:anchor-subsolution}
 \int_Dw_\eps\nabla v\cdot\nabla\psi\dd w
 =-\int_{\partial A\cap D}w_\eps\psi\,\partial_{n_A}v\dd\sigma
 \le0.
\end{equation}
Here $\dd\sigma$ is surface measure, interpreted as counting measure
when $d=1$; the boundary term on $\partial D$ is zero because
$\psi$ has compact support in $D$.
Fix $y$ with $B_{2\eps}(y)\subset D$ and take $\eps$ small enough
that this ball is contained in a flat coordinate chart. Let
\begin{equation}\label{eq:rescaled-weight}
 \widehat v(\xi):=v(y+\eps\xi),\qquad
 a_{\eps,y}(\xi):=\frac{w_\eps(y+\eps\xi)}{w_\eps(y)}
 =\exp\!\left(-\frac{L(y+\eps\xi)-L(y)}\eps\right),
 \quad \xi\in B_2.
\end{equation}
The fundamental theorem of calculus gives, for $\xi\in B_2$,
\[
 |L(y+\eps\xi)-L(y)|
 =\left|\eps\int_0^1\nabla L(y+t\eps\xi)\cdot\xi\dd t\right|
 \le2\eps\|\nabla L\|_\infty.
\]
Consequently, with $\lambda_0:=e^{-2\|\nabla L\|_\infty}$ and
$\Lambda_0:=e^{2\|\nabla L\|_\infty}$,
\begin{equation}\label{eq:rescaled-ellipticity}
 0<\lambda_0\le a_{\eps,y}(\xi)\le\Lambda_0,
 \qquad \xi\in B_2.
\end{equation}
Changing variables in \eqref{eq:anchor-subsolution} gives
\begin{equation}\label{eq:rescaled-subsolution}
 \int_{B_2}a_{\eps,y}\nabla\widehat v\cdot\nabla\psi\dd\xi\le0,
 \qquad 0\le\psi\in C_c^\infty(B_2).
\end{equation}
We derive the following local bound with a constant depending only
on $d,\lambda_0,\Lambda_0$:
\begin{equation}\label{eq:local-mean}
 \sup_{w\in B_\eps(y)}v(w)\le C\eps^{-d/2}
 \|v\|_{L^2(B_{2\eps}(y))},\qquad B_{2\eps}(y)\subset D.
\end{equation}
Let $p\ge2$, $1\le r<R\le2$, and choose
$\eta\in C_c^\infty(B_R)$ with $0\le\eta\le1$, equal to $1$ on $B_r$, and
$|\nabla\eta|\le C/(R-r)$. For $\delta\in(0,1)$ let
$v_\delta:=\widehat v+\delta$. Since $0\le\widehat v\le1$,
we may extend $t\mapsto(t+\delta)^{p-1}$ from $[0,1]$ to a
$C^1(\R)$ function with bounded derivative. The Sobolev chain rule
\cite[Proposition 3.21(2)]{Hunter2014}, followed by the product rule,
gives
\[
 \nabla\left(\eta^2v_\delta^{p-1}\right)
 =2\eta v_\delta^{p-1}\nabla\eta
  +(p-1)\eta^2v_\delta^{p-2}\nabla\widehat v\in L^2(B_R;\R^d).
\]
Let $\varphi:=\eta^2v_\delta^{p-1}$ and $K:=\operatorname{supp}\eta$.
Since $K$ is compactly contained in $B_R$, the zero extension
$\widetilde\varphi$ belongs to $H^1(\R^d)$ and its weak gradient is
the zero extension of $\nabla\varphi$. Indeed, for
$\psi\in C_c^\infty(\R^d)$, testing the weak derivative of
$\varphi$ with $\psi$ times a cutoff equal to one near $K$ gives
\[
 \int_{\R^d}\widetilde\varphi\,\partial_j\psi\dd\xi
 =\int_{B_R}\varphi\,\partial_j\psi\dd\xi
 =-\int_{B_R}\partial_j\varphi\,\psi\dd\xi,
 \qquad 1\le j\le d.
\]
Choose $\varrho\in C_c^\infty(B_1)$ with $\varrho\ge0$ and
$\int_{\R^d}\varrho(\xi)\dd\xi=1$. For
$0<\sigma<\dist(K,\partial B_R)$, define
\[
 \varrho_\sigma(\xi):=\sigma^{-d}\varrho(\xi/\sigma),\qquad
 \varphi_\sigma(x):=\int_{\R^d}
          \varrho_\sigma(x-\xi)\widetilde\varphi(\xi)\dd\xi.
\]
Then $\varphi_\sigma\ge0$ and
$\operatorname{supp}\varphi_\sigma\subset K+\overline{B_\sigma}
\subset B_R$, so $\varphi_\sigma\in C_c^\infty(B_R)$.
Minkowski's integral inequality and translation continuity in $L^2(\R^d)$, applied to
$\widetilde\varphi$ and its first derivatives, gives
\[
 \left\|\varphi_\sigma-\widetilde\varphi\right\|_{H^1(\R^d)}
 \le\int_{B_1}\varrho(\xi)
    \left\|\widetilde\varphi(\cdot-\sigma\xi)-\widetilde\varphi\right\|_{H^1(\R^d)}
       \dd\xi\longrightarrow0.
\]
In particular, $\varphi\in H_0^1(B_R)$. Apply
\eqref{eq:rescaled-subsolution} to $\varphi_\sigma$ and use
\eqref{eq:rescaled-ellipticity} to pass to the limit, since
\[
 \left|\int_{B_R}a_{\eps,y}\nabla\widehat v\cdot
             \nabla(\varphi_\sigma-\varphi)\dd\xi\right|
 \le\Lambda_0\|\nabla\widehat v\|_{L^2(B_R)}
             \left\|\nabla(\varphi_\sigma-\varphi)\right\|_{L^2(B_R)}
 \longrightarrow0.
\]
The Cauchy--Schwarz inequality and Young's inequality give
\begin{align}
 (p-1)\int_{B_R}a_{\eps,y}\eta^2v_\delta^{p-2}
                 |\nabla\widehat v|^2\dd\xi
 &\le2\int_{B_R}a_{\eps,y}\eta v_\delta^{p-1}
                 |\nabla\widehat v|\,|\nabla\eta|\dd\xi,
 \notag\\
 \int_{B_R}\eta^2v_\delta^{p-2}|\nabla\widehat v|^2\dd\xi
 &\le\frac{4\Lambda_0^2}{(p-1)^2\lambda_0^2}
       \int_{B_R}|\nabla\eta|^2v_\delta^p\dd\xi.
 \label{eq:moser-test}
\end{align}
Expanding $\nabla\left(\eta v_\delta^{p/2}\right)$ and using
\eqref{eq:moser-test} gives
\[
 \int_{B_R}\left|\nabla\left(\eta v_\delta^{p/2}\right)\right|^2\dd\xi
 \le Cp^2\int_{B_R}|\nabla\eta|^2v_\delta^p\dd\xi.
\]
For fixed $p$, boundedness of $\widehat v$ and
$\nabla\widehat v\in L^2(B_R)$ justify passage to
$\delta\downarrow0$ by the dominated convergence theorem. Thus,
\begin{equation}\label{eq:moser-energy}
 \int_{B_R}\left|\nabla\left(\eta\widehat v^{p/2}\right)\right|^2\dd\xi
 \le Cp^2\int_{B_R}|\nabla\eta|^2\widehat v^p\dd\xi.
\end{equation}
For $f\in H_0^1(B_2)$, extended by zero, the Sobolev inequality used
here is
\begin{equation}\label{eq:unit-sobolev}
 \|f\|_{L^{2\chi}(B_2)}^2\le C_d\|\nabla f\|_{L^2(B_2)}^2,
 \qquad
 \chi=\begin{cases}d/(d-2),&d>2,\\2,&d\le2.\end{cases}
\end{equation}
For $d>2$, this is the Gagliardo--Nirenberg--Sobolev inequality
\cite[Section 5.6.1, Theorem 1]{Evans2010}, with exponent $2$.
For $d=2$, the same theorem with exponent $4/3$, followed by
H\"older's inequality on $B_2$, gives
\[
 \|f\|_{L^4(B_2)}\le C\|\nabla f\|_{L^{4/3}(B_2)}
 \le C|B_2|^{1/4}\|\nabla f\|_{L^2(B_2)}.
\]
For $d=1$, $f(t)=\int_{-2}^t f'(s)\dd s$ gives
$\|f\|_\infty\le2\|f'\|_2$ and hence \eqref{eq:unit-sobolev}.
Applying \eqref{eq:unit-sobolev} to
$f=\eta\widehat v^{p/2}$ and using \eqref{eq:moser-energy} yields
\begin{equation}\label{eq:moser-step}
 \|\widehat v\|_{L^{p\chi}(B_r)}
 \le\left(\frac{Cp^2}{(R-r)^2}\right)^{1/p}
       \|\widehat v\|_{L^p(B_R)},\qquad 1\le r<R\le2.
\end{equation}
Let $p_j:=2\chi^j$ and $R_j:=1+2^{-j}$. Iterating
\eqref{eq:moser-step} with $R=R_j$ and $r=R_{j+1}$ gives
\[
 \|\widehat v\|_{L^{p_{n+1}}(B_{R_{n+1}})}
 \le\prod_{j=0}^n\left(Cp_j^2 4^{j+1}\right)^{1/p_j}
           \|\widehat v\|_{L^2(B_2)}.
\]
The product is bounded independently of $n$, since
\[
 \sum_{j=0}^\infty
 \frac{\log C+2\log p_j+(j+1)\log4}{p_j}
 \le C'\sum_{j=0}^\infty\frac{j+1}{\chi^j}<\infty.
\]
Letting $n\to\infty$ and changing variables in
\eqref{eq:rescaled-weight}, we obtain
\[
 \sup_{w\in B_\eps(y)}v(w)
 =\|\widehat v\|_{L^\infty(B_1)}
 \le C\|\widehat v\|_{L^2(B_2)}
 =C\eps^{-d/2}\|v\|_{L^2(B_{2\eps}(y))},
\]
which is \eqref{eq:local-mean}. For
$2\eps<\dist(\overline{D_0},\T^d\setminus D)$, apply
\eqref{eq:local-mean} at each $y\in D_0$ and then
\eqref{eq:anchor-l2} to prove \eqref{eq:anchor-sup}.
Finally, the change of variable $f\mapsto1-f$ in \eqref{eq:cap} and
uniqueness of the minimizer of $f\mapsto\cE_\eps(f,f)$ over
$\mathcal A_{B,A}$ give
\[
 q_{B,A}=1-q_{A,B},\qquad
 \capc_\eps(B,A)=\capc_\eps(A,B).
\]
Apply \eqref{eq:anchor-l2}--\eqref{eq:anchor-sup} to $q_{B,A}$
with anchor $S_B$ to obtain
\eqref{eq:zero-anchor-l2}--\eqref{eq:zero-anchor-sup}.
\end{proof}

The next estimate makes a small-sphere average of the mean time
asymptotically identical to its value at the sphere's center. A lift by
one coordinate reduces the Poisson equation to a homogeneous equation.

\begin{lemma}[Relative flatness at scale $\eps^2$]\label{lem:flat}
Let $u_\eps(x):=\E_x[\tau_B]$. For every compact
$F\subset\T^d\setminus B$, there are $C<\infty$ and $\alpha>0$ such that
\begin{equation}\label{eq:flat}
 \sup_{y\in\overline{B_{\eps^2}(x)}}|u_\eps(y)-u_\eps(x)|
 \le C\eps^\alpha u_\eps(x),\qquad x\in F,
\end{equation}
for all sufficiently small $\eps$.
\end{lemma}

\begin{proof}
Fix $\eps>0$. Using the quotient map $\mathfrak p$ from Section~\ref{sec:notation}, define the periodic function
$\widetilde L(\xi):=L(\mathfrak p(\xi))$ on $\R^d$. The periodic
lift of \eqref{eq:sde} is the Euclidean diffusion
\begin{equation}\label{eq:periodic-lift}
 \mathrm d\widetilde X_t=-\nabla\widetilde L(\widetilde X_t)\dd t
                  +\sqrt{2\eps}\,\mathrm dB_t,\qquad
 \widetilde{\cL}_\eps=\eps\Delta-\nabla\widetilde L\cdot\nabla,
 \qquad X_t=\mathfrak p(\widetilde X_t),
\end{equation}
where $B_t$ is standard Brownian motion on $\R^d$ and
$\mathfrak p(\widetilde X_0)=X_0$. Periodicity and $L\in C^3$
give bounded derivatives of $\widetilde L$ through order three.
In particular, the drift in \eqref{eq:periodic-lift} is bounded
and globally Lipschitz, and its principal coefficient matrix is
$\eps I_d$.
Let $\widetilde p_\eps(t,\xi,\zeta)$ be the transition density
of \eqref{eq:periodic-lift}. As a function of $(t,\zeta)$, it is
the fundamental solution of the forward equation
\begin{equation}\label{eq:lifted-forward-equation}
 \partial_t\widetilde p_\eps
 =\eps\Delta_\zeta\widetilde p_\eps
  +\operatorname{div}_\zeta
       \left(\widetilde p_\eps\nabla\widetilde L(\zeta)\right),\qquad
 \widetilde p_\eps(0,\xi,\cdot)=\delta_\xi.
\end{equation}
The initial condition is in the sense of distributions.
The principal matrix $\eps I_d$ is uniformly elliptic for this
fixed $\eps$, and the coefficients are bounded and continuously
differentiable. The fundamental-solution bounds in
\cite[Theorems 7 and 10(ii)]{Aronson1968} apply to
\eqref{eq:lifted-forward-equation}: for each $T>0$ there is
$C=C(\eps,T,L,d)\ge1$ such that
\begin{equation}\label{eq:lifted-gaussian-bounds}
 \frac{e^{-C|\xi-\zeta|^2/t}}{Ct^{d/2}}
 \le\widetilde p_\eps(t,\xi,\zeta)
 \le\frac{C}{t^{d/2}}e^{-|\xi-\zeta|^2/(Ct)},
 \qquad 0<t\le T,\quad \xi,\zeta\in\R^d.
\end{equation}
In \eqref{eq:lifted-forward-equation}, the principal coefficient
matrix is $\eps I_d$, the first-order coefficients are
$\partial_j\widetilde L$, $1\le j\le d$, and the zeroth-order
coefficient is $\Delta\widetilde L$. These coefficients are bounded
and continuously differentiable because $L\in C^3$ is periodic.
For these coefficients, the fundamental solution
$\widetilde p_\eps(t,\xi,\zeta)$ is continuous in $(t,\xi,\zeta)$
for $t>0$; see the classical-solution construction in
\cite[Section 7, pp.\ 658--660]{Aronson1968}.
Its projection to the torus has density
\begin{equation}\label{eq:periodized-density}
 p_\eps(t,x,y):=\sum_{k\in\mathbb Z^d}
       \widetilde p_\eps(t,\widetilde x,\widetilde y+2\pi k),
 \qquad \mathfrak p(\widetilde x)=x,\quad
        \mathfrak p(\widetilde y)=y.
\end{equation}
Periodicity of the coefficients in \eqref{eq:periodic-lift} makes
\eqref{eq:periodized-density} independent of the representatives.
The upper bound in \eqref{eq:lifted-gaussian-bounds} makes the series
locally uniformly convergent for $t>0$, and the lower bound makes
every summand positive. Thus $p_\eps$ is continuous and strictly
positive. Projecting the law of $\widetilde X_t$ gives
$\mathbb P_x(X_t\in E)=\int_Ep_\eps(t,x,y)\dd y$ for Borel
$E\subset\T^d$.
Choose a closed ball $B_0\subset\Int B$ of positive volume and fix
$t_0>0$. The set $\T^d\times B_0$ is compact, and the function
$(x,y)\mapsto p_\eps(t_0,x,y)$ is continuous and strictly positive
there. Thus its minimum is attained and is positive:
\begin{equation}\label{eq:hitting-minorization}
 m_\eps:=\min_{(x,y)\in\T^d\times B_0}p_\eps(t_0,x,y)>0,
 \qquad p_\eps^*:=m_\eps|B_0|>0.
\end{equation}
For every $x\in\T^d$,
\[
 \mathbb P_x(\tau_B\le t_0)
 \ge\mathbb P_x(X_{t_0}\in B_0)
 =\int_{B_0}p_\eps(t_0,x,y)\dd y\ge p_\eps^*.
\]
Applying the Markov property at the times $nt_0$ therefore yields
\begin{align}
 \mathbb P_x(\tau_B>(n+1)t_0)
 &\le(1-p_\eps^*)\mathbb P_x(\tau_B>nt_0)
 \le(1-p_\eps^*)^{n+1},\qquad n\ge0,\notag\\
 \E_x[\tau_B]
 &=\int_0^\infty\mathbb P_x(\tau_B>t)\dd t
 \le t_0\sum_{n=0}^\infty(1-p_\eps^*)^n
 =\frac{t_0}{p_\eps^*}<\infty.
 \label{eq:hitting-finite}
\end{align}
The constants $m_\eps$ and $p_\eps^*$ in
\eqref{eq:hitting-minorization} may depend on $\eps$.
Let $\Omega:=\T^d\setminus B$. We show that $u_\eps$ is the unique
weak solution in $H_0^1(\Omega)$, and also the unique classical solution
in $C^2(\Omega)\cap C(\overline\Omega)$, of
\begin{equation}\label{eq:poisson}
 -\cL_\eps u_\eps=1\quad\text{on }\Omega,
 \qquad u_\eps|_{\partial\Omega}=0.
\end{equation}
The extension to $B$ is defined to be zero. Since
$\cL_\eps f=\eps w_\eps^{-1}\operatorname{div}(w_\eps\nabla f)$,
the weak formulation of \eqref{eq:poisson} is
\begin{equation}\label{eq:poisson-weak}
 \eps\int_\Omega w_\eps\nabla v_\eps\cdot\nabla\psi\dd w
 =\int_\Omega w_\eps\psi\dd w,
 \qquad \psi\in H_0^1(\Omega).
\end{equation}
To obtain a solution, extend $f\in H_0^1(\Omega)$ by zero to $B$.
The argument for \eqref{eq:torus-anchored-poincare}, now with
zero anchor $S=B$, gives
\[
 \|f\|_{L^2(\Omega)}\le P_B\|\nabla f\|_{L^2(\Omega)},\qquad
 P_B:=1+\left(\frac{\Vol(\T^d)}{|B|}\right)^{1/2}.
\]
Consequently,
\begin{equation}\label{eq:poisson-coercivity}
 \eps\int_\Omega w_\eps|\nabla f|^2\dd w
 \ge\frac{\eps e^{-\max_{\T^d}L/\eps}}{1+P_B^2}
       \|f\|_{H^1(\Omega)}^2.
\end{equation}
Minimizing
\[
 J_\eps(f):=\frac\eps2\int_\Omega w_\eps|\nabla f|^2\dd w
                    -\int_\Omega w_\eps f\dd w,
 \qquad f\in H_0^1(\Omega),
\]
therefore gives a weak solution $v_\eps$ of
\eqref{eq:poisson-weak}: \eqref{eq:poisson-coercivity} makes
$J_\eps$ coercive, and $J_\eps$ is weakly lower semicontinuous.
For two weak solutions $v_\eps^{(1)},v_\eps^{(2)}$, subtract
\eqref{eq:poisson-weak} and test with their difference to obtain
\[
 \eps\int_\Omega w_\eps
       |\nabla(v_\eps^{(1)}-v_\eps^{(2)})|^2\dd w=0.
\]
Equation \eqref{eq:poisson-coercivity} then proves uniqueness.

For fixed $\eps$, the matrix $\eps w_\eps I_d$ is uniformly elliptic,
its coefficients and $w_\eps$ are $C^2$, and $\partial\Omega$ is smooth.
The local boundary and interior regularity for weak solutions in
\cite[Corollary 8.36]{GilbargTrudinger2001}, applied in finitely many
flat charts, gives $v_\eps\in C^{1,\beta}(\overline\Omega)$ for
$0<\beta<1$. In an interior ball, the equation becomes
\[
 \Delta v_\eps=\eps^{-1}
       \left(\nabla L\cdot\nabla v_\eps-1\right),
\]
whose right-hand side is H\"older continuous. Local Dirichlet
solvability in \cite[Theorem 6.13]{GilbargTrudinger2001} and uniqueness
of the weak Dirichlet solution on that ball give
$v_\eps\in C^2(\Omega)$.
Applying Dynkin's formula to \eqref{eq:poisson} for $v_\eps$,
first on compact subdomains of $\Omega$ and then letting them
increase to $\Omega$, yields
\[
 v_\eps(x)=\E_x[t\wedge\tau_B]
       +\E_x\!\left[v_\eps(X_t)\mathbf1_{\{\tau_B>t\}}\right],
 \qquad x\in\Omega.
\]
Continuity and the zero boundary values justify this passage.
The last expectation has absolute value at most
$\|v_\eps\|_\infty\mathbb P_x(\tau_B>t)\to0$ by
\eqref{eq:hitting-finite}. Letting $t\to\infty$ proves
$v_\eps(x)=\E_x[\tau_B]=u_\eps(x)>0$ on $\Omega$.
Positivity follows from path continuity and $\dist(x,B)>0$.
The same stopped identity applies to any
$C^2(\Omega)\cap C(\overline\Omega)$ solution of
\eqref{eq:poisson}, proving uniqueness in that class as well.

For $x\in F$, choose a representative $\widetilde x\in\R^d$ and
use the chart $\xi\mapsto\mathfrak p(\widetilde x+\xi)$, whose
metric tensor is $I_d$.
Choose $\eps_0\in(0,1)$ so that
$8\eps_0<\min\{\dist(F,B),\pi\}$, and define on $B_8^{d+1}$,
for $0<\eps<\eps_0$,
\[
 W_{\eps,x}(y,t):=u_\eps(x+\eps y)+\frac{\eps t^2}{2}.
\]
Equation \eqref{eq:poisson} implies the exact homogeneous equation
\begin{equation}\label{eq:lift}
 \left(\Delta_y+\partial_t^2
 -\nabla L(x+\eps y)\cdot\nabla_y\right)W_{\eps,x}=0.
\end{equation}
The choice of $\eps_0$ ensures that \eqref{eq:lift} holds
on $B_8^{d+1}$ for every $x\in F$ and $0<\eps<\eps_0$.
Its principal coefficient matrix is $I_{d+1}$, and its drift
$(-\nabla L(x+\eps y),0)$ has norm at most
$\|\nabla L\|_\infty$ there. Harnack's inequality and the interior
H\"older estimate in \cite[Lemmas 4.1 and 4.2]{BEGK2004}, applied
to the positive solution $W_{\eps,x}$ of \eqref{eq:lift}, give
\begin{align}
 \sup_{(y,t)\in B_2^{d+1}}W_{\eps,x}(y,t)
 &\le C_H W_{\eps,x}(0,0),\notag\\
 \osc_{(y,t)\in B_\eps^{d+1}}W_{\eps,x}(y,t)
 &\le C_{\mathrm{Hol}}\eps^\alpha
       \sup_{(y,t)\in B_2^{d+1}}W_{\eps,x}(y,t)
 \le C_{\mathrm{Hol}}C_H\eps^\alpha W_{\eps,x}(0,0).
 \label{eq:lift-regularity}
\end{align}
The constants $C_H,C_{\mathrm{Hol}}$ and $\alpha\in(0,1)$ depend
only on $d$ and $\|\nabla L\|_\infty$; they are independent of
$x\in F$ and $0<\eps<\eps_0$. The threshold $\eps_0$ also depends
on $\dist(F,B)$. For $y\in\overline{B_{\eps^2}(x)}$, the point
$((y-x)/\eps,0)$ lies in $\overline{B_\eps^{d+1}}$.
Thus \eqref{eq:lift-regularity} and continuity give \eqref{eq:flat}
with $C=C_{\mathrm{Hol}}C_H$.
\end{proof}

\begin{proof}[Proof of Theorem~\ref{thm:reduction}]
Fix a nonempty compact set $Q\subset U$.
A connected smooth domain $D$ with closure in $U$
can be chosen to contain $A_*$ and $Q$ in its interior. Indeed, finitely
many balls cover the compact sets, and finitely many paths inside the
open connected set $U$ join these balls. Their union has a connected
smooth neighborhood $D$ with
\begin{equation}\label{eq:source-domains}
 A_*\cup Q\subset D,\qquad \overline D\subset D',\qquad
 \overline{D'}\subset U,
\end{equation}
where $D'$ is a slightly larger connected smooth domain.
All closures in \eqref{eq:source-domains} are compact; in particular,
they avoid $B$. Let $b:=\max_{w\in\overline{D'}}L(w)<h$.

First enlarge the source set from $A_*$ to $\overline D$. Apply
Lemma~\ref{lem:anchor} to $q_*=q_{A_*,B}$, with a fixed positive-volume
subset of $A_*$ as anchor. By \eqref{eq:cap-assumption},
\begin{equation}\label{eq:source-enlargement}
 \sup_{w\in\overline D}(1-q_*(w))\le
 C\eps^{-N}e^{-(h-b)/(2\eps)}=:\eta_\eps=o(1),
\end{equation}
for some fixed $N$. The function
$\min\{q_* /(1-\eta_\eps),1\}$ is admissible for
$(\overline D,B)$. Monotonicity and \eqref{eq:energy} therefore give
\begin{equation}\label{eq:enlarged-cap}
 \capc_\eps(A_*,B)\le\capc_\eps(\overline D,B)
 \le(1-\eta_\eps)^{-2}\capc_\eps(A_*,B).
\end{equation}

For $x\in Q$, let $A_{\eps,x}:=\overline{B_{\eps^2}(x)}$ and
$q_x:=q_{A_{\eps,x},B}$. For small $\eps$, these balls lie in $D$.
By \eqref{eq:enlarged-cap},
\begin{equation}\label{eq:small-cap-upper}
 \capc_\eps(A_{\eps,x},B)
 \le (1+o(1))\capc_\eps(A_*,B)
 \le C\eps^{d/2}e^{-h/\eps},
\end{equation}
uniformly over $x\in Q$. Apply Lemma~\ref{lem:anchor} in $D'$ to $q_x$,
with $S=A_{\eps,x}$, whose volume is $\omega_d\eps^{2d}$.
Equations \eqref{eq:anchor-sup} and \eqref{eq:small-cap-upper} imply
\begin{equation}\label{eq:small-q}
 \sup_{x\in Q}\sup_{w\in\overline D}(1-q_x(w))
 \le C\eps^{-N'}e^{-(h-b)/(2\eps)}=:r_\eps=o(1).
\end{equation}
Consequently $\min\{q_x/(1-r_\eps),1\}$ is admissible for $(A_*,B)$,
so
\begin{equation}\label{eq:small-cap}
 (1-r_\eps)^2\capc_\eps(A_*,B)
 \le\capc_\eps(A_{\eps,x},B)
 \le(1+o(1))\capc_\eps(A_*,B).
\end{equation}

The compact set $\T^d\setminus(D\cup\Int B)$ contains no zero of $L$.
Thus $L\ge c>0$ there. Since $q_x=0$ on $B$, \eqref{eq:small-q} gives
\begin{equation}\label{eq:numerator}
 \left|\int_{\T^d\setminus B}q_x w_\eps\dd w
       -\int_Dw_\eps\dd w\right|
 \le r_\eps\Vol(D)+\Vol(\T^d)e^{-c/\eps}
 =o(M_A(\eps)),
\end{equation}
uniformly over $x\in Q$. Lemma~\ref{lem:mass} applies to $D$.

We derive the equilibrium-measure identity
\eqref{eq:equilibrium-measure}, corresponding to
\cite[equation (2.27)]{BEGK2004}. In
$\Omega:=\T^d\setminus B$, define the equilibrium measure by
\begin{equation}\label{eq:equilibrium-distribution}
 -\eps\operatorname{div}(w_\eps\nabla q_x)=\nu_x
 \quad\text{in }\mathcal D'(\Omega),\qquad
 \int_\Omega\varphi\dd\nu_x
 =\eps\int_\Omega w_\eps\nabla q_x\cdot\nabla\varphi\dd w,
 \quad\varphi\in C_c^\infty(\Omega).
\end{equation}
With $n_x$ the outward unit normal to $A_{\eps,x}$, its
boundary density is
\[
 \nu_x(\mathrm dw)=-\eps w_\eps(w)\partial_{n_x}q_x(w)\,
                  \mathrm d\sigma(w),\qquad w\in\partial A_{\eps,x}.
\]
Since $q_x=1$ on $A_{\eps,x}$ and $0\le q_x\le1$, this density
is nonnegative. Let $n_B$ be the outward unit normal to $B$,
which points into $\Omega$.
For $\varphi\in\{q_x,u_\eps\}$, integration by parts on
$\Omega\setminus A_{\eps,x}$ gives
\begin{align}
 \eps\int_\Omega w_\eps\nabla q_x\cdot\nabla\varphi\dd w
 &=-\eps\int_{\partial A_{\eps,x}}
           \varphi w_\eps\partial_{n_x}q_x\dd\sigma
   -\eps\int_{\partial B}
           \varphi w_\eps\partial_{n_B}q_x\dd\sigma\notag\\
 &=\int_{\partial A_{\eps,x}}\varphi\dd\nu_x,
 \label{eq:equilibrium-boundary-test}
\end{align}
because both test functions vanish on $\partial B$.
Thus \eqref{eq:equilibrium-distribution} holds also for these
two test functions. Taking $\varphi=q_x$ in
\eqref{eq:equilibrium-boundary-test} gives
the total mass of $\nu_x$ because $q_x=1$ on its support.
Testing it against $u_\eps$, and testing \eqref{eq:poisson} in
weighted divergence form against $q_x$, gives
\begin{equation}\label{eq:equilibrium-measure}
 \nu_x(\partial A_{\eps,x})=\capc_\eps(A_{\eps,x},B),\qquad
 \int_{\partial A_{\eps,x}}u_\eps\dd\nu_x
 =\cE_\eps(q_x,u_\eps)
 =\int_\Omega q_x w_\eps\dd w.
\end{equation}
The factor $w_\eps$ is included in our measure $\nu_x$.

By the first identity in \eqref{eq:equilibrium-measure},
\[
 \widehat\nu_x:=\frac{\nu_x}{\capc_\eps(A_{\eps,x},B)}
\]
is a probability measure on $\partial A_{\eps,x}$. Dividing the
second identity in \eqref{eq:equilibrium-measure} by
$\capc_\eps(A_{\eps,x},B)$ gives
\begin{equation}\label{eq:normalized-equilibrium}
 \int_{\partial A_{\eps,x}}u_\eps(y)\widehat\nu_x(\mathrm dy)
 =\frac{\displaystyle\int_\Omega q_xw_\eps\dd w}
        {\capc_\eps(A_{\eps,x},B)}.
\end{equation}
Applying \eqref{eq:flat} with $F=Q$ yields
\begin{align*}
 \left|\int_{\partial A_{\eps,x}}u_\eps(y)\widehat\nu_x(\mathrm dy)
             -u_\eps(x)\right|
 &\le\sup_{y\in\partial A_{\eps,x}}|u_\eps(y)-u_\eps(x)|\le C\eps^\alpha u_\eps(x),\qquad x\in Q.
\end{align*}
Consequently \eqref{eq:normalized-equilibrium} implies
\begin{align*}
 \E_x[\tau_B]
 =\frac{\displaystyle\int_\Omega q_xw_\eps\dd w}
         {\capc_\eps(A_{\eps,x},B)}(1+\mathcal{O}(\eps^\alpha))=\frac{M_A(\eps)}{\capc_\eps(A_*,B)}(1+o(1)),\qquad x\in Q,
\end{align*}
where the second equality uses \eqref{eq:small-cap} and
\eqref{eq:numerator}, with errors uniform over $Q$. This is
\eqref{eq:reduction-conclusion}.
\end{proof}

The basin restriction in Theorem~\ref{thm:main} follows from
connectedness of a finite gradient-trajectory segment.
Since $K_A$ is compact and contained in $U$, there is $\delta>0$
such that $\{w:\dist(w,K_A)<\delta\}\subset U$. If
$x\in\mathcal B_A$ and $L(x)<h$, then $\phi_T(x)\in U$ for some
finite $T$, and
\[
 L(\phi_t(x))=L(x)-\int_0^t|\nabla L(\phi_s(x))|^2\dd s<h,
 \qquad 0\le t\le T.
\]
Thus the trajectory segment joins $x$ to $U$ within $\{L<h\}$,
so that $x\in U$.

The smoothness imposed on $B$ in the proof can be removed for the
distance neighborhoods in the problem. This also explains why the
leading constant does not depend on a sufficiently small fixed radius.

\begin{corollary}[Changing the target neighborhood]\label{cor:target}
Under Theorem~\ref{thm:reduction}, let $T$ be any closed neighborhood of
$K_B$ with $T\subset U_B$. Then \eqref{eq:reduction-conclusion} holds
with $\tau_T$ in place of $\tau_B$, retaining the original fixed
capacity $\capc_\eps(A_*,B)$.
\end{corollary}

\begin{proof}
Choose smooth closed neighborhoods $B_-$ and $B_+$ of $K_B$ such that
\[
 B_-\subset \Int B\cap\Int T,\qquad
 B\cup T\subset\Int B_+,\qquad B_+\subset U_B.
\]
Such sets exist because both compact sets lie in the open connected
region $U_B$.
Apply the zero-anchored estimate \eqref{eq:zero-anchor-sup} of
Lemma~\ref{lem:anchor} to $q=q_{A_*,B}$ on a smooth connected
domain compactly contained in $U_B$ and containing $B_+$ in its
interior. By \eqref{eq:cap-assumption}, there are $N<\infty$ and
$\delta>0$ such that
\[
 \sup_{w\in B_+}q(w)\le C\eps^{-N}e^{-\delta/(2\eps)}=:s_\eps=o(1).
\]
The function $(q-s_\eps)_+/(1-s_\eps)$ is admissible for
$(A_*,B_+)$. Monotonicity and \eqref{eq:cap} give
\[
 \capc_\eps(A_*,B)\le\capc_\eps(A_*,B_+)
 \le(1-s_\eps)^{-2}\capc_\eps(A_*,B).
\]
To compare the capacity for $B$ with that for the smaller target
$B_-\subset B$, monotonicity of \eqref{eq:cap} and
\eqref{eq:cap-assumption} give
\begin{equation}\label{eq:target-shrink-upper}
 \capc_\eps(A_*,B_-)\le\capc_\eps(A_*,B)
 \le C\eps^{d/2}e^{-h/\eps}.
\end{equation}
Using \eqref{eq:target-shrink-upper}, apply \eqref{eq:zero-anchor-sup} to
$p=q_{A_*,B_-}$ on a connected region compactly contained in $U_B$
and containing $B$. Its anchor has fixed positive volume. Thus
$\sup_{w\in B}p(w)\le\widetilde s_\eps=o(1)$, and
$(p-\widetilde s_\eps)_+/(1-\widetilde s_\eps)$ is admissible
for $(A_*,B)$. It follows that
\[
 \left(1-\widetilde s_\eps\right)^2\capc_\eps(A_*,B)
 \le\capc_\eps(A_*,B_-)\le\capc_\eps(A_*,B).
\]
Theorem~\ref{thm:reduction} applies to both $B_-$ and $B_+$ with the
same $c_*$. Finally $\tau_{B_+}\le\tau_T\le\tau_{B_-}$, which proves
the conclusion uniformly over each $Q$.
\end{proof}

\subsection{Global localization at the unique Morse gate}\label{sec:general}

\subsubsection{Analytic topology and boundary gates}

The link between the gate definition and the local saddle geometry
uses boundary accessibility. We record the exact topological facts,
including existence of a minimizing path, before constructing a gate
on the boundary of each strict sublevel component.

\begin{lemma}[Topology of the critical sublevel set]\label{lem:tame-topology}
The following facts hold on the compact analytic torus.
\begin{enumerate}[label=(\roman*),leftmargin=2em]
 \item A compact semianalytic set has finitely many connected
       components, each compact and path connected. It is locally
       path connected. Connected components and closures of
       semianalytic sets are semianalytic.
 \item If $S$ is semianalytic and $p\in\overline S\setminus S$, there
       is a continuous path $\alpha\colon[0,1]\to\overline S$ such
       that $\alpha(0)=p$ and $\alpha(t)\in S$ for $0<t\le1$.
 \item The family $\mathcal P_h$ in Definition~\ref{def:gate} is
       nonempty.
\end{enumerate}
\end{lemma}

\begin{proof}
\textit{(i) Components and local path connectedness.}
Corollaries 2.7 and 2.8 of \cite{BierstoneMilman1988} show that the
components of a semianalytic set are semianalytic and locally finite,
and that its closure is semianalytic. For a compact semianalytic set
$X$, its pairwise disjoint connected components satisfy
\begin{equation}\label{eq:tame-components}
 X=\bigcup_{j=1}^N C_j,
 \qquad C_j\text{ compact, connected, and semianalytic},\qquad N<\infty.
\end{equation}
Let $\iota\colon\T^d\to\R^{2d}$ be the analytic embedding
\[
 \iota(w_1,\ldots,w_d)
   :=(\cos w_1,\sin w_1,\ldots,\cos w_d,\sin w_d).
\]
For each $C_j$, the set $\iota(C_j)$ is compact, connected, and
subanalytic: in a finite family of relatively compact coordinate
charts it is the projection of the semianalytic graph of $\iota$.
Recall that a continuous curve $\alpha\colon[0,1]\to\R^{2d}$ is
\emph{rectifiable} if its length
\[
 \ell(\alpha):=\sup_{N\ge1,\,0=t_0<t_1<\cdots<t_N=1}
       \sum_{k=1}^N|\alpha(t_k)-\alpha(t_{k-1})|
\]
is finite; see \cite[Definition 2.3.1]{BuragoBuragoIvanov2001}.
Theorem 6.10 of \cite{BierstoneMilman1988} therefore joins any
$\iota(a),\iota(b)\in\iota(C_j)$ by a rectifiable curve
$\widetilde\gamma\colon[0,1]\to\iota(C_j)$.
The path
$\gamma:=\iota^{-1}\circ\widetilde\gamma$ lies in $C_j$ and joins
$a$ to $b$.

For $p\in X$ and a sufficiently small coordinate radius $r>0$,
apply \eqref{eq:tame-components} to $X\cap\overline{B_r(p)}$ to obtain
pairwise disjoint components $C_{r,j}$ such that
\[
 X\cap\overline{B_r(p)}=\bigcup_{j=1}^{N(r)} C_{r,j},
 \qquad p\in C_{r,1}.
\]
If $N(r)>1$, choose
\[
 0<\delta<\min\left\{r,
       \min_{2\le j\le N(r)}\dist(p,C_{r,j})\right\};
\]
if $N(r)=1$, choose any $\delta\in(0,r)$. Compactness and
disjointness of the components make this choice possible, and
\begin{equation}\label{eq:tame-local-neighborhood}
 X\cap B_\delta(p)\subset C_{r,1}\subset X\cap\overline{B_r(p)}.
\end{equation}
The set $C_{r,1}$ is path connected by the preceding argument.
As $r$ can be arbitrarily small, \eqref{eq:tame-local-neighborhood}
provides a basis of path-connected neighborhoods of $p$ in $X$.

\textit{(ii) Accessibility from a semianalytic set.}
The rectilinearization theorem
\cite[Theorem 0.2]{BierstoneMilman1988}, applied near $p$, provides
finitely many analytic maps $\phi_i\colon\R^d\to\T^d$ and compact
sets $F_i\subset\R^d$, $1\le i\le J$, such that
\begin{equation}\label{eq:rectilinear-cover}
 N_p\subset\bigcup_{i=1}^J\phi_i(F_i),\qquad
 \phi_i^{-1}(S)=\bigcup_{Q\in\mathcal Q_i}Q,
\end{equation}
where $N_p$ is a neighborhood of $p$ and $\mathcal Q_i$ is a finite
family of coordinate quadrants. Choose $p_n\in S\cap N_p$ with
$p_n\to p$. By \eqref{eq:rectilinear-cover}, each $p_n$ has a lift
$v_n\in F_{i_n}\cap Q_n$, with $\phi_{i_n}(v_n)=p_n$.
Finiteness of the indices and quadrants, followed by compactness of
$F_i$, gives a subsequence, still indexed by $n$, satisfying
\begin{equation}\label{eq:rectilinear-lifts}
 i_n=i,\qquad Q_n=Q,\qquad
 v_n\longrightarrow v\in F_i\cap\overline Q,\qquad \phi_i(v)=p.
\end{equation}
Fix $v'\in Q$ and let $v(t):=(1-t)v+tv'$ for $0\le t\le1$.
For each coordinate whose sign in $Q$ is
$\sigma_k\in\{-1,1\}$,
\[
 \sigma_k v_k(t)=(1-t)\sigma_k v_k+t\sigma_k v'_k>0,
 \qquad 0<t\le1,
\]
since $v\in\overline Q$ and $v'\in Q$. Coordinates prescribed to
be zero remain zero. Thus $v(t)\in Q$ for $t>0$, and
\begin{equation}\label{eq:accessible-arc}
 \alpha(t):=\phi_i(v(t)),\qquad
 \alpha(0)=p,\qquad \alpha((0,1])\subset S.
\end{equation}
Continuity and \eqref{eq:rectilinear-lifts} prove (ii).

\textit{(iii) Existence of a minimizing path.}
For $n\ge1$, let $X_n$ be the component of
$\{L\le h+1/n\}$ containing $K_A$. By the definition of $h$,
there is a connecting path $\gamma_n$ such that
$\max_{t\in[0,1]}L(\gamma_n(t))<h+1/n$.
The union $K_A\cup\gamma_n([0,1])\cup K_B$ is connected, and hence
\begin{equation}\label{eq:nested-sublevels}
 K_A\cup K_B\subset X_{n+1}\subset X_n,\qquad
 X_n\subset\{L\le h+1/n\}.
\end{equation}
Each $X_n$ is compact and connected. Their intersection
$X_\infty:=\bigcap_{n=1}^\infty X_n$ is compact and connected.
Indeed, if it were the union of two disjoint nonempty compact sets,
choose disjoint open neighborhoods $O_1,O_2$ of these sets.
Nested compactness implies $X_n\subset O_1\cup O_2$ for some $n$:
otherwise a convergent subsequence of points in
$X_n\setminus(O_1\cup O_2)$ would lie in
$X_\infty\setminus(O_1\cup O_2)$. Since each $X_n$ meets both
$O_1$ and $O_2$, this would contradict connectedness of $X_n$.
By \eqref{eq:nested-sublevels},
\[
 K_A\cup K_B\subset X_\infty\subset X:=\{L\le h\}.
\]
Thus $X_\infty$ lies in one connected component $C$ of the compact
semianalytic set $X$. Part (i), applied to $C$, provides a path
$\gamma$ from $K_A$ to $K_B$ within $C$. Consequently,
\[
 h\le\max_{t\in[0,1]}L(\gamma(t))\le h,
 \qquad\text{and hence }\gamma\in\mathcal P_h.
\]
This completes the proof.
\end{proof}

The next lemma constructs a gate that is minimal even among arbitrary
subsets of the communication level. It connects the gate definition
of Bovier, Eckhoff, Gayrard, and Klein
\cite[Definition 1.1]{BEGK2004}
to the strict sublevel components.

\begin{lemma}[A gate on each well boundary]\label{lem:boundary-gate}
Let $X=\{L\le h\}$ and let $C_B$ be the connected component of
$X\setminus\overline U$ containing $K_B$. Then
\begin{equation}\label{eq:boundary-gate}
 G_A:=\partial_X C_B\subset\partial U\subset\{L=h\}
\end{equation}
is a nonempty closed gate. More precisely, it meets every minimizing
path, and for every $p\in G_A$ there is a minimizing path $\gamma_p$
such that
\begin{equation}\label{eq:gate-witness}
 \gamma_p([0,1])\cap G_A=\{p\}.
\end{equation}
Interchanging $A$ and $B$ gives a gate $G_B\subset\partial U_B$
with the same properties.
\end{lemma}

\begin{proof}
The boundary of $U$ is contained in $\{L=h\}$. Indeed, continuity
gives $L\le h$ there, and any boundary point with $L<h$ would have
a connected open neighborhood in the same strict sublevel component.
In particular $K_B\cap\overline U=\varnothing$.

By Lemma~\ref{lem:tame-topology} and
\cite[Corollaries 2.7 and 2.8]{BierstoneMilman1988},
$Y=X\setminus\overline U$ is semianalytic with finitely many
semianalytic connected components. These components are closed in
$Y$ and, because there are finitely many, also open in $Y$.
Thus $C_B$ is open in $X$ and contains the connected set $K_B$.
It is path connected, since $X$ is locally path connected and
$C_B$ is connected and open in $X$.

The relative closure satisfies
$\overline{C_B}^{\,X}\cap Y=C_B$. Since $C_B$ is open in $X$,
\[
 G_A=\overline{C_B}^{\,X}\setminus C_B\subset\overline U.
\]
Moreover $U$ is open and disjoint from $C_B$, so
$U\cap\overline{C_B}^{\,X}=\varnothing$. This proves
\eqref{eq:boundary-gate}; it also shows that $G_A$ is closed in the
compact set $X$. Every minimizing path lies in $X$, starts in $U$,
and ends in $C_B$. It therefore crosses $\partial_X C_B=G_A$.
Lemma~\ref{lem:tame-topology}(iii) provides at least one minimizing
path. Since that path meets $G_A$, we have $G_A\ne\varnothing$.

Fix $p\in G_A$. Then $p\in\overline U\setminus U$ and
$p\in\overline{C_B}\setminus C_B$. Part (ii) of
Lemma~\ref{lem:tame-topology}, applied separately to the semianalytic
sets $U$ and $C_B$, gives two arcs ending at $p$, with all their other
points in $U$ and $C_B$, respectively. The open connected set $U$
contains a path from $K_A$ to the first arc; path connectedness of
$C_B$ gives a path from the second arc to $K_B$. Concatenating these
paths and arcs gives a path $\gamma_p$ in $X$ from $K_A$ to $K_B$.
Because $U\cap G_A=C_B\cap G_A=\varnothing$, it satisfies
\eqref{eq:gate-witness}. Its maximum loss is $h$, since $L(p)=h$.
Removing any $p$ from $G_A$ thus destroys the level-cut property.
Every proper subset omits such a point, so $G_A$ is a gate in the
sense of Definition~\ref{def:gate}, as well as a minimal closed cut.
The argument with the two wells interchanged proves the last assertion.
\end{proof}

\subsubsection{Coordinates and the two sides of the gate}

Exact quadratic coordinates allow the saddle neighborhood to remain
fixed while $\eps\downarrow0$. Their metric error disappears in a
second limit as the neighborhood shrinks.

\begin{lemma}[Quadratic coordinates with the correct metric]
\label{lem:morse-chart}
Choose orthonormal coordinates centered at $z$ in which
$H_z=\operatorname{diag}(-\mu_-,H_+)$, with $H_+$ positive definite.
There is an analytic coordinate map $w=\Phi(s,y)$ with
$\Phi(0)=z$ and $D\Phi(0)=I$ such that
\begin{equation}\label{eq:morse-exact}
 L(\Phi(s,y))=h-\frac{\mu_-s^2}{2}+\frac{y^{\top}H_+y}{2}.
\end{equation}
If $\widetilde f=f\circ\Phi$, its energy on a coordinate region $E$
is
\begin{equation}\label{eq:chart-energy}
 \cE_{\eps,\Phi(E)}(f,f)
 =\eps\int_E\left(\nabla\widetilde f\right)^{\top}\mathsf G(q)
       \nabla\widetilde f\,e^{-L(\Phi(q))/\eps}\dd q,
 \qquad
 \mathsf G=|\det D\Phi|(D\Phi)^{-1}(D\Phi)^{-\top}.
\end{equation}
Here $q=(s,y)$ and $\mathsf G(0)=I$. On any family of coordinate boxes
shrinking to zero, there is $\omega(r)\downarrow0$ such that
\begin{equation}\label{eq:metric-bound}
 (1-\omega(r))I\preceq\mathsf G(q)\preceq(1+\omega(r))I.
\end{equation}
\end{lemma}

\begin{proof}
We give a construction to retain the derivative normalization.
Choose a representative $\widetilde z\in\R^d$ and an orthogonal
matrix $O$ whose columns are eigenvectors of $H_z$, with the
negative direction first. The map
$\Psi_z(x):=\mathfrak p(\widetilde z+Ox)$ is a flat chart centered
at $z$ with Euclidean metric tensor. Let
$\ell(x):=L(\Psi_z(x))$ and
$H:=\nabla^2\ell(0)=\operatorname{diag}(-\mu_-,H_+)$.
Since $\ell(0)=h$ and $\nabla\ell(0)=0$, Taylor's formula gives
\begin{equation}\label{eq:morse-taylor}
 \ell(x)-h=\frac12x^{\top}A(x)x,\qquad
 A(x)=2\int_0^1(1-t)\nabla^2\ell(tx)\dd t,
 \qquad A(0)=H.
\end{equation}
For small $x$, the binomial power series defines the real analytic
matrix $T(x)=(H^{-1}A(x))^{1/2}$ near the identity. Since
$(H^{-1}A)^{\top}H=H(H^{-1}A)$, the same power-series identity gives
$T^{\top}H=HT$. Therefore,
\[
 T(x)^{\top}HT(x)=H(T(x))^2=A(x),\qquad
 \ell(x)-h=\frac12(T(x)x)^{\top}H(T(x)x).
\]
The analytic map $x\mapsto T(x)x$ has derivative $I$ at zero; its
local analytic inverse $\psi$ therefore satisfies $D\psi(0)=I$.
Let $\Phi:=\Psi_z\circ\psi$. In the chosen orthonormal frame,
$D\Phi(0)=I$, and \eqref{eq:morse-taylor} proves \eqref{eq:morse-exact}.
The change-of-variables formula and the chain rule give
\eqref{eq:chart-energy}; continuity of $D\Phi$ gives
\eqref{eq:metric-bound}.
\end{proof}

\Needspace{14\baselineskip}
The single-gate condition contains both a separation statement and a
connection statement. The latter ensures that the two low faces of
the quadratic neighborhood belong to the prescribed wells.

\begin{lemma}[Equivalent gate criteria]\label{lem:gate-criterion}
For an index-one Morse critical point $z$ at height $h$, the following
conditions are equivalent:
\begin{enumerate}[label=(\roman*),leftmargin=2em]
 \item $\{z\}$ is the unique gate of Definition~\ref{def:gate}.
 \item $\{z\}$ is the unique inclusion-minimal closed level cut.
 \item Every $\gamma\in\mathcal P_h$ meets $z$, and some
       $\gamma_0\in\mathcal P_h$ satisfies
       $\gamma_0([0,1])\cap\{L=h\}=\{z\}$.
\end{enumerate}
Under these conditions, $z\in\partial U\cap\partial U_B$. After
possibly reversing $s$ in \eqref{eq:morse-exact}, the local descending
lobe with $s<0$ lies in $U$, and the one with $s>0$ lies in $U_B$.
\end{lemma}

\begin{proof}
Assume either (i) or (ii). The sets $G_A\subset\partial U$ and
$G_B\subset\partial U_B$ from Lemma~\ref{lem:boundary-gate} are
both gates and minimal closed level cuts. Uniqueness in either
class therefore gives
\begin{equation}\label{eq:singleton-boundary-gates}
 G_A=G_B=\{z\},\qquad z\in\partial U\cap\partial U_B.
\end{equation}
Also, $\{z\}$ is a level cut, so every minimizing path meets $z$.

In the coordinates of Lemma~\ref{lem:morse-chart}, choose a small
box $\mathcal N=(-a,a)\times B_b^{d-1}$ in the chart. By
\eqref{eq:morse-exact}, its strict sublevel set is the disjoint union
\begin{equation}\label{eq:descending-lobes}
 \begin{split}
 \{L<h\}\cap\Phi(\mathcal N)&=\mathcal L_-\cup\mathcal L_+,\\
 \mathcal L_\pm&:=\Phi\left\{(s,y)\in\mathcal N:
                 y^{\top}H_+y<\mu_-s^2,\ \pm s>0\right\}.
 \end{split}
\end{equation}
For $d=1$, omit $y$ and use the two half-intervals.
Each lobe is connected:
one can first shrink $y$ to zero and then move along its half-axis.
By \eqref{eq:singleton-boundary-gates}, every neighborhood of $z$
meets both $U$ and $U_B$. Equation~\eqref{eq:descending-lobes}
therefore places points of these two components in the two lobes.
A connected lobe lies in a single component of $\{L<h\}$, and
$U\ne U_B$, so the two components meet different lobes.
Reversing $s$ if necessary gives
\begin{equation}\label{eq:lobe-membership}
 \mathcal L_-\subset U,\qquad \mathcal L_+\subset U_B.
\end{equation}
Connect a point of $K_A$ to the negative half-axis inside the open
connected set $U$, cross the axis through $z$, and connect the positive
half-axis to $K_B$ inside $U_B$. The two outer paths have compact images
strictly below $h$, and the middle segment reaches $h$ only at $z$.
Their concatenation is the path required in (iii).

Conversely, under (iii), every level cut meets $\gamma_0$ at $z$
and therefore contains $z$. Since $\{z\}$ itself meets every minimizing
path, it is the unique gate. It is closed, so it is also the unique
minimal closed level cut. This proves both (i) and (ii).

Applying the implication from (i) already proved above gives
\eqref{eq:singleton-boundary-gates} and \eqref{eq:lobe-membership}
under (iii) as well, establishing the final assertions of the lemma
under each of the equivalent conditions.
\end{proof}

If uniqueness is formulated for the whole critical level, the gate
condition follows directly by decreasing the loss along its gradient
flow.

\begin{corollary}[A sufficient critical-level condition]
\label{cor:critical-level}
Suppose $\mathcal P_h\ne\varnothing$ and
\[
 \{w:L(w)=h,\ \nabla L(w)=0\}=\{z\},
\]
where $z$ is a nondegenerate index-one point. Then
Assumption~\ref{ass:single-gate} holds.
\end{corollary}

\begin{proof}
Let $\phi_t$ be the negative gradient flow and fix $t>0$. Along it,
\[
 L(\phi_t(w))=L(w)-\int_0^t|\nabla L(\phi_s(w))|^2\dd s.
\]
Every point of $\{L=h\}\setminus\{z\}$ therefore moves strictly
below $h$, whereas $z$ and the zero set are fixed. For any
$\gamma\in\mathcal P_h$ avoiding $z$, the compact path
$\phi_t\circ\gamma$ would have maximum strictly below $h$, a
contradiction. For any minimizing path $\gamma$, the new path
$\phi_t\circ\gamma$ meets $\{L=h\}$ only at $z$. 
By applying
Lemma~\ref{lem:gate-criterion}, we complete the proof.
\end{proof}

\subsubsection{A separating function outside a fixed saddle box}

Lemma~\ref{lem:tame-topology} provides the finite-component and
path-connectedness properties needed to separate the wells outside
the saddle neighborhood.
Suppose Assumption~\ref{ass:single-gate} holds, and let $A_*\subset U$ and
$B\subset U_B$ be fixed closed neighborhoods with smooth boundaries.
Choose the orientation from
Lemma~\ref{lem:gate-criterion}. For $d\ge2$, fix $R>1$ such that
$\lambda_{\min}(H_+)R^2>4\mu_-$ and set
\begin{align}
 \mathcal C_r:=[-r,r]\times\overline{B_{Rr}^{d-1}},\qquad
 D_r:=\Phi(\mathcal C_r),\qquad
 \mathcal C_r^{\rm out}:=(-2r,2r)\times B_{2Rr}^{d-1}.
 \label{eq:boxes}
\end{align}
Take $r$ small enough that the outer box is in the chart and its closure
avoids $A_*\cup B$. In dimension one, omit the $y$ coordinates and
use the corresponding intervals.

The following construction provides the global part of the capacity
upper bound. Its positive energy gap may depend on $r$, which is held
fixed throughout the small-temperature limit.

\begin{lemma}[Global extension of the saddle profile]\label{lem:separator}
For each sufficiently small fixed $r$, define
\begin{equation}\label{eq:general-profile}
 g_\eps(s):=
 \begin{cases}
  1,&s\le-r,\\
  I_\eps^{-1}\displaystyle\int_s^r
                   e^{-\mu_-t^2/(2\eps)}\dd t,&-r<s<r,\\
  0,&s\ge r,
 \end{cases}
 \qquad I_\eps:=\int_{-r}^r e^{-\mu_-t^2/(2\eps)}\dd t.
\end{equation}
There is an admissible $f_\eps\in H^1(\T^d)$ for $(A_*,B)$ such that
$f_\eps(\Phi(s,y))=g_\eps(s)$ on $\mathcal C_r$ and
\begin{equation}\label{eq:outside-energy}
 \cE_{\eps,\T^d\setminus D_r}(f_\eps,f_\eps)
 \le C_r(1+\eps)e^{-(h+\eta_r)/\eps},
\end{equation}
for some $\eta_r>0$ and $C_r<\infty$, independent of small $\eps$.
\end{lemma}

\begin{proof}
The set
\[
 S_r=\{L\le h\}\setminus\Int D_r
\]
is compact and semianalytic.
Lemma~\ref{lem:tame-topology}(i), specifically
\eqref{eq:tame-components}, gives pairwise disjoint components
$S_{r,j}$ satisfying
\begin{equation}\label{eq:separator-components}
 S_r=\bigcup_{j=1}^{N_r}S_{r,j},\qquad N_r<\infty,
 \qquad S_{r,j}\text{ compact and path connected}.
\end{equation}
The intersections of $S_r$ with the two end faces of $D_r$ are
\begin{equation}\label{eq:separator-end-sections}
 \begin{split}
 E_- =\Phi\left\{(-r,y):y^{\top}H_+y\le\mu_-r^2\right\},\qquad
 E_+ =\Phi\left\{(r,y):y^{\top}H_+y\le\mu_-r^2\right\}.
 \end{split}
\end{equation}
For $d\ge2$, the lateral boundary is
\begin{equation}\label{eq:lateral-boundary}
 \partial_{\mathrm{lat}}D_r
 :=\Phi\left([-r,r]\times\partial B_{Rr}^{d-1}\right).
\end{equation}
For a point $\Phi(s,y)$ of \eqref{eq:lateral-boundary},
\eqref{eq:morse-exact} and the choice of $R$ imply
\[
 L(\Phi(s,y))-h
 \ge\tfrac12\left(\lambda_{\min}(H_+)R^2-\mu_-\right)r^2>0.
\]
Thus $S_r\cap\partial_{\mathrm{lat}}D_r=\varnothing$.
For $d=1$ this boundary is empty and $E_\pm=\{\Phi(\pm r)\}$.
Every point of $E_-$ can be joined to $K_A$ in
$\{L\le h\}\setminus\{z\}$: first shrink $y$ on the face to zero,
then use the negative lobe and a path in $U$. Every point of $E_+$
has the analogous connection to $K_B$.

No component $S_{r,j}$ in \eqref{eq:separator-components} can meet
both $A_*\cup E_-$ and $B\cup E_+$. Otherwise a path in $S_{r,j}$,
concatenated with the connections to $K_A$ and $K_B$ just constructed,
would give a path $\gamma$ with
\[
 \gamma(0)\in K_A,\qquad \gamma(1)\in K_B,\qquad
 \gamma([0,1])\subset\{L\le h\}\setminus\{z\}.
\]
Its maximum loss is $h$ by the definition of the communication
height, contradicting Lemma~\ref{lem:gate-criterion}(iii).
Let
\[
 J_1:=\{j\in\{1,\ldots,N_r\}:S_{r,j}\cap(A_*\cup E_-)\ne\varnothing\},
 \quad
 F_1:=\bigcup_{j\in J_1}S_{r,j},\quad
 F_0:=\bigcup_{j\in\{1,\ldots,N_r\}\setminus J_1}S_{r,j}.
\]
By \eqref{eq:separator-components}, $F_1$ and $F_0$ are disjoint
compact sets, with $A_*\cup E_-\subset F_1$ and
$B\cup E_+\subset F_0$.
Choose $v\in C^\infty(\T^d;[0,1])$ equal to $1$ on a neighborhood of
$F_1$ and $0$ on a neighborhood of $F_0$.

Let $0\le\chi\le1$ be a smooth cutoff equal to $1$ on a neighborhood
of $D_r$, supported in $\Phi(\mathcal C_r^{\rm out})$. Within the chart set
\begin{equation}\label{eq:glued-test}
 f_\eps=v+\chi(g_\eps\circ s-v),
\end{equation}
and set $f_\eps=v$ outside the chart. This is an admissible $H^1$
function, and it equals $g_\eps(s)$ on $D_r$.

We check that its exterior gradient vanishes near $S_r$, uniformly
in $\eps$. In the outer box, a point outside $D_r$ with $L\le h$
must have $|s|\ge r$. Indeed, if $|y|\ge Rr$ and $|s|\le2r$, then
\begin{equation}\label{eq:lateral-gap}
 L(\Phi(s,y))-h
 \ge\frac12\left(\lambda_{\min}(H_+)R^2-4\mu_-\right)r^2>0.
\end{equation}
When $s\le-r$, such a point connects inside $S_r$ to $E_-$ by
shrinking $y$ and moving along the negative half-axis; its label is
$1$. When $s\ge r$, the corresponding label is $0$. Hence $v$ and
$g_\eps$ have the same constant value near these points, on the
exterior side of $D_r$. Away from the cutoff support, $\nabla f_\eps
=\nabla v=0$ near $S_r$. At an end face this statement still holds
in a relative neighborhood outside $D_r$, since its low-energy part
is separated from the lateral face by \eqref{eq:lateral-gap}.

Compactness of $S_r$ provides a fixed relatively open neighborhood
$\mathcal N_r$ of $S_r$ in $\T^d\setminus\Int D_r$ on which
$\nabla f_\eps=0$ almost everywhere for every $\eps$. Define
\begin{equation}\label{eq:exterior-compact-complement}
 F_r:=(\T^d\setminus\Int D_r)\setminus\mathcal N_r.
\end{equation}
The set $F_r$ is compact and disjoint from $\{L\le h\}$, so there
is $\eta_r>0$ with $L\ge h+\eta_r$ on $F_r$.
Equation \eqref{eq:general-profile} gives
$|g_\eps'|\le C_r\eps^{-1/2}$ for small $\eps$; differentiation of
\eqref{eq:glued-test} then yields
\[
 |\nabla f_\eps|\le C_r\left(1+\eps^{-1/2}\right).
\]
By \eqref{eq:energy} and \eqref{eq:exterior-compact-complement},
\begin{align*}
 \cE_{\eps,\T^d\setminus D_r}(f_\eps,f_\eps)
 &=\eps\int_{F_r}|\nabla f_\eps|^2e^{-L/\eps}\dd w\\
 &\le C_r\eps(1+\eps^{-1/2})^2\Vol(F_r)e^{-(h+\eta_r)/\eps}\\
 &\le C_r(1+\eps)e^{-(h+\eta_r)/\eps},
\end{align*}
which proves \eqref{eq:outside-energy}.
For $d=1$ the lateral region is
absent and the same argument uses only the two end points.
\end{proof}

\subsubsection{Capacity asymptotics and the transition law}

The upper bound now has its entire leading contribution inside the
saddle box. The anchored estimate from Section~\ref{sec:reduction}
provides the matching boundary values for a lower bound.

\begin{theorem}[Capacity for arbitrary analytic wells]
\label{thm:general-capacity}
Under Assumption~\ref{ass:single-gate}, for any fixed smooth closed
neighborhoods $A_*\subset U$ and $B\subset U_B$ of the two zero
components,
\begin{equation}\label{eq:general-capacity}
 \capc_\eps(A_*,B)
 =c_z\eps^{d/2}e^{-h/\eps}(1+o(1)),
\end{equation}
where $c_z$ is defined in \eqref{eq:cz}.
\end{theorem}

\begin{proof}
Fix a small $r$ and use $f_\eps$ from Lemma~\ref{lem:separator}.
The Dirichlet principle, \eqref{eq:chart-energy}, and
\eqref{eq:metric-bound} give
\begin{align}
 \capc_\eps(A_*,B)
 &\le (1+\omega(r))\eps e^{-h/\eps}
 \frac{\displaystyle\int_{|y|\le Rr}
                 e^{-y^{\top}H_+y/(2\eps)}\dd y}{I_\eps}
 +C_r(1+\eps)e^{-(h+\eta_r)/\eps},
 \label{eq:general-upper}
\end{align}
where
\[
I_\eps^{-1}=\int_{-r}^r |g_\eps'(s)|^2 e^{\mu_-s^2/(2\eps)}\dd s.
\]
For every fixed $r$ the Gaussian integrals satisfy
\begin{equation}\label{eq:gaussian-limits}
 I_\eps\sim\sqrt{\frac{2\pi\eps}{\mu_-}},\qquad
 \int_{|y|\le Rr}e^{-y^{\top}H_+y/(2\eps)}\dd y
 \sim\frac{(2\pi\eps)^{(d-1)/2}}{\sqrt{\det H_+}}.
\end{equation}
In particular,
\begin{equation}\label{eq:general-limsup}
 \limsup_{\eps\downarrow0}
 \frac{\capc_\eps(A_*,B)}{\eps^{d/2}e^{-h/\eps}}
 \le(1+\omega(r))
       \frac{\sqrt{\mu_-}(2\pi)^{d/2-1}}{\sqrt{\det H_+}}.
\end{equation}

For the lower bound in dimension $d\ge2$, choose $\beta>0$ with
$\lambda_{\max}(H_+)\beta^2<\mu_-/2$ and restrict to
$\widehat{\mathcal C}_r=[-r,r]\times\overline{B_{\beta r}^{d-1}}$.
Define its two end faces in the torus by
\begin{equation}\label{eq:lower-end-faces}
 F_{r,-}:=\Phi\left(\{-r\}\times\overline{B_{\beta r}^{d-1}}\right),
 \qquad
 F_{r,+}:=\Phi\left(\{r\}\times\overline{B_{\beta r}^{d-1}}\right).
\end{equation}
For $|y|\le\beta r$, \eqref{eq:morse-exact} gives
\[
 L(\Phi(\pm r,y))
 \le h-\tfrac12\left(\mu_--\lambda_{\max}(H_+)\beta^2\right)r^2
 <h-\tfrac14\mu_-r^2.
\]
Consequently the faces in \eqref{eq:lower-end-faces} lie strictly
below $h$, and \eqref{eq:lobe-membership} gives
$F_{r,-}\subset U$ and $F_{r,+}\subset U_B$. Each compact face,
together with a fixed positive-volume subset of its well neighborhood,
is contained in a connected smooth region compactly contained in the
appropriate strict sublevel component. Let the maximum loss on these
two regions be at most $h-\delta_r$, with $\delta_r>0$.

Write $q=q_{A_*,B}$ and $\widetilde q=q\circ\Phi$.
Equation \eqref{eq:general-upper} gives
$\capc_\eps(A_*,B)\le C_r\eps^{d/2}e^{-h/\eps}$.
Lemma~\ref{lem:anchor}, with fixed positive-volume anchors, therefore
implies
\begin{equation}\label{eq:general-face}
 \sup_{|y|\le\beta r}
 \left(\left|1-\widetilde q(-r,y)\right|+\left|\widetilde q(r,y)\right|\right)
 \le C_r\eps^{-N}e^{-\delta_r/(2\eps)}=:e_{r,\eps}=o(1).
\end{equation}
For each $y$, the fundamental theorem of calculus and the weighted
Cauchy--Schwarz inequality give
\[
 \int_{-r}^r|\partial_s\widetilde q(s,y)|^2
                         e^{\mu_-s^2/(2\eps)}\dd s
 \ge\frac{[\widetilde q(-r,y)-\widetilde q(r,y)]^2}{I_\eps}.
\]
Dropping the stable derivative terms in the Euclidean lower bound
from \eqref{eq:metric-bound}, and using \eqref{eq:general-face}, yields
\begin{equation}\label{eq:general-lower}
 \capc_\eps(A_*,B)
 \ge(1-\omega(r))(1-e_{r,\eps})^2\eps e^{-h/\eps}
 \frac{\displaystyle\int_{|y|\le\beta r}
                  e^{-y^{\top}H_+y/(2\eps)}\dd y}{I_\eps}.
\end{equation}
The Gaussian limits \eqref{eq:gaussian-limits}, with $\beta r$ in
place of $Rr$, give the lower counterpart to
\eqref{eq:general-limsup}, with coefficient $1-\omega(r)$.
Since $|\det H_z|=\mu_-\det H_+$, the common limiting coefficient
as $r\downarrow0$ is exactly $c_z$. The order of limits is
$\eps\downarrow0$ for each fixed $r$, followed by $r\downarrow0$.
For each fixed $r$, the gaps $\eta_r$ and $\delta_r$ are positive
constants throughout the first limit.
For $d=1$, there are no stable coordinates $y$. The Gaussian
integrals over $y$ in \eqref{eq:general-upper},
\eqref{eq:gaussian-limits}, and \eqref{eq:general-lower} are then
integrals over $\R^0=\{0\}$, with measure $1$ and quadratic form $0$,
and hence equal $1$. With the convention $\det H_+=1$, the two
faces in \eqref{eq:lower-end-faces} become the interval end points
$\{\Phi(-r)\}$ and $\{\Phi(r)\}$. The anchored estimates and the
one-dimensional Cauchy--Schwarz inequality above give the same
upper and lower bounds.
\end{proof}

\begin{proof}[Proof of Theorem~\ref{thm:main}]
Choose fixed smooth closed well neighborhoods $A_*\subset U$ and
$B\subset U_B$. Theorem~\ref{thm:general-capacity} proves the
assumption of Theorem~\ref{thm:reduction} with $c_*=c_z$.
Lemma~\ref{lem:mass} identifies $c_A=a_A\Gamma(\lambda_A+1)$.
For small fixed $\rho$, the compact distance neighborhood $B_\rho$
is contained in $U_B$; Corollary~\ref{cor:target} then gives
\eqref{eq:main-uniform}. The identity $C_A=c_A/c_z$ is
\eqref{eq:main-constant}. The gradient-trajectory argument after
Theorem~\ref{thm:reduction} verifies the required basin restriction.
\end{proof}

\section{Conclusion}\label{sec:conclusion}\label{sec:scope-checks}

We establish an Eyring--Kramers formula for a low-temperature Langevin
diffusion between two compact connected zero components of a
real-analytic potential.
Our main result relates the mean transition time from an
analytic singular well to its sublevel-volume data and the Hessian
of a unique Morse gate. Analytic topology provides the global
separation needed for capacity localization, and the anchored
energy estimate gives uniformity for deterministic initial states.
In particular, through the ratio of
well mass to saddle capacity, the equilibrium scaling
$\eps^{\lambda_A}(\log(1/\eps))^{m_A-1}$ determines the
subexponential factor of a mean transition time, where the subscript $A$
refers to the source well around the zero component $K_A$ in
\eqref{eq:zeros}, the parameters $\lambda_A$ and $m_A$ are,
respectively, the exponent and multiplicity in the sublevel-volume
asymptotic \eqref{eq:volume} for a fixed neighborhood $U_A$ of $K_A$.
The explicit example shows that a logarithmic correction can arise
from a loss of transverse nondegeneracy along a smooth set of minima.

For singular learning theory, the Eyring--Kramers formula gives a dynamical role to
the learning coefficient and its multiplicity. 
Equation~\eqref{eq:log-transition} shows how wells with the same
barrier height can have different persistence times because of
their singular geometry. In the AI safety motivation, this provides
a quantitative component of a theory of behavioral stability under
continued noisy optimization: once a learning model's distinct
behaviors have been associated with wells satisfying the stated
assumptions, the theorem determines their mean transition time.
Applications to delayed changes in generalization additionally
require a link between these wells and test performance, while
finite-horizon transition probabilities require information about
the hitting-time distribution.

In the future, it will be interesting to investigate
the case of several competing gates and
degenerate transition saddles. Several gates require a global
description of how their local capacities combine; a degenerate
saddle requires a capacity calculation adapted to its unstable
and stable directions. Quantitative remainder estimates would
also be useful, particularly when logarithmic corrections make
the leading asymptotic converge slowly.
For learning dynamics, future directions include state-dependent SGD
noise and approximation estimates valid on the metastable time
scale. Extending the argument to noncompact parameter spaces requires
return-time and tail estimates that replace the compactness arguments
used here.

\bibliographystyle{alpha}
\bibliography{bibtex}

\appendix

\section{An Analytic Well with a Logarithmic Correction}\label{sec:model}

\subsection{Well geometry and the unique gate}

We consider a concrete potential and verify every ingredient for this example. Let
$\eta=1/2$ and define on $\T^2$
\begin{align}
 a(x)&:=1-\cos x,\qquad
 V(x):=(1-\cos x)^2(1+\cos x)e^{\eta\sin x},
 \label{eq:model-components}\\
 L(x,y)&:=V(x)+a(x)(1-\cos y).
 \label{eq:model}
\end{align}
This is a nonnegative real-analytic periodic function.
Its zero components and local volume coefficient can be computed
without resolution of singularities.

\begin{proposition}[Well geometry and volume]\label{prop:model-mass}
For \eqref{eq:model},
\begin{equation}\label{eq:model-zeros}
 K_A=\{0\}\times\T,\qquad K_B=\{(\pi,0)\}.
\end{equation}
For a sufficiently small fixed strip $U_A=\{|x|<a_0\}$,
\begin{equation}\label{eq:model-volume}
 \Vol\{(x,y)\in U_A:L(x,y)<s\}
 \sim 2\sqrt{s}\log(1/s),\qquad s\downarrow0.
\end{equation}
Consequently,
\begin{equation}\label{eq:model-mass}
 \lambda_A=\frac12,\quad m_A=2,\quad a_A=2,\quad
 M_A(\eps)\sim\sqrt{\pi}\sqrt{\eps}\log(1/\eps),\quad\eps\downarrow0.
\end{equation}
\end{proposition}

\begin{proof}
The zeros of $V$ are $0$ and $\pi$ modulo $2\pi$. Since $a(0)=0$
and $a(\pi)=2$, \eqref{eq:model-zeros} follows from the nonnegativity
of both terms in \eqref{eq:model}. At $(0,0)$, Taylor expansion gives
\begin{equation}\label{eq:model-local}
 L(x,y)=\frac{x^4}{2}+\frac{x^2y^2}{4}
 +\mathcal{O}\left(|x|^5+x^4y^2+x^2y^4\right).
\end{equation}
For $y\ne0$ on $K_A$, the transverse Hessian is
$\partial_x^2L(0,y)=1-\cos y>0$. At $(0,0)$ it vanishes.

We compute the volume for
$P(x,y)=\alpha x^4+\beta x^2y^2$ on $[-r,r]^2$, with
$\alpha,\beta>0$ and fixed $r>0$, as $s\downarrow0$. For small $s$, let
$x_1=(s/\alpha)^{1/4}$ and let $x_0>0$ solve
$\alpha x_0^4+\beta r^2x_0^2=s$. Then
$x_0=\sqrt{s}/(\sqrt{\beta}r)(1+\mathcal{O}(s))$. Define
\begin{equation}\label{eq:model-integral-correction}
 R_s:=\int_{x_0/x_1}^1\frac{\sqrt{1-u^4}-1}{u}\dd u.
\end{equation}
The substitution $u=x/x_1$ gives
\begin{align}
 \int_{x_0}^{x_1}\frac{\sqrt{s-\alpha x^4}}{x}\dd x
 &=\sqrt{s}\left(\log\frac{x_1}{x_0}+R_s\right),
 \label{eq:model-integral-split}\\
 |R_s|&\le\int_0^1\frac{1-\sqrt{1-u^4}}{u}\dd u
 =\int_0^1\frac{u^3}{1+\sqrt{1-u^4}}\dd u
 \le\frac14.
 \label{eq:model-correction-bound}
\end{align}
Moreover,
\[
 \log\frac{x_1}{x_0}
 =\frac14\log(1/s)+\log\left(\sqrt{\beta}r\alpha^{-1/4}\right)
  +\mathcal O(s),\qquad s\downarrow0.
\]
Using \eqref{eq:model-integral-split} and
\eqref{eq:model-correction-bound}, we obtain
\begin{align}
 \Vol\left\{(x,y)\in[-r,r]^2:P(x,y)<s\right\}
 &=4rx_0+\frac4{\sqrt{\beta}}
       \int_{x_0}^{x_1}\frac{\sqrt{s-\alpha x^4}}{x}\dd x
 \notag\\
 &=4rx_0+\frac4{\sqrt{\beta}}\sqrt{s}
       \left(\log\frac{x_1}{x_0}+R_s\right)
 \notag\\
 &=\frac1{\sqrt{\beta}}\sqrt{s}\log(1/s)
       +\mathcal{O}(\sqrt{s}),\qquad s\downarrow0.
 \label{eq:model-poly-volume}
\end{align}
In particular, the correction to the logarithmic integral is
$\sqrt{s}R_s$, whose absolute value is at most $\sqrt{s}/4$ by
\eqref{eq:model-correction-bound}. The constants in the
$\mathcal O(\sqrt{s})$ term in \eqref{eq:model-poly-volume} may
depend on $\alpha,\beta,r$.

By \eqref{eq:model-local}, for each $\delta\in(0,1)$ there is a small square
on which $(1-\delta)P\le L\le(1+\delta)P$, with
$\alpha=1/2$ and $\beta=1/4$. Equation \eqref{eq:model-poly-volume}
squeezes the leading coefficient between $2/\sqrt{1+\delta}$ and
$2/\sqrt{1-\delta}$. Outside this square, the portion with $|x|\ge r$
has $L\ge V\ge c_r>0$ and makes no contribution for small $s$.
On the portion with $|y|\ge r$, the inequality $L(x,y)\ge c x^2$
gives sublevel volume $\mathcal{O}(\sqrt{s})$. Letting $\delta\downarrow0$ proves
\eqref{eq:model-volume}. Lemma~\ref{lem:mass} and
$2\Gamma(3/2)=\sqrt{\pi}$ prove \eqref{eq:model-mass}.
\end{proof}

The asymmetry in $e^{\eta\sin x}$ separates the two possible routes
around the $x$-circle. The next proposition verifies
Assumption~\ref{ass:single-gate} by checking both path conditions
of Lemma~\ref{lem:gate-criterion} and the Hessian at the gate.

\begin{proposition}[Unique Morse gate]\label{prop:gate}
Let $t_-\in(0,\pi)$ be the unique solution of
\begin{equation}\label{eq:troot}
 2\cot t_-+\cot(t_-/2)-\eta\cos t_-=0,
 \qquad \eta=\frac12.
\end{equation}
The communication height for \eqref{eq:model} is
\begin{equation}\label{eq:h-model}
 h=(1-\cos t_-)^2(1+\cos t_-)e^{-\eta\sin t_-},
\end{equation}
and the unique minimizing gate is $\{z\}$ with $z=(-t_-,0)$.
In particular, Assumption~\ref{ass:single-gate} holds.
The Hessian at $z$ is
$\operatorname{diag}(-\mu,\kappa)$, where
\begin{equation}\label{eq:model-hessian}
 \kappa=1-\cos t_-,\qquad
 \mu=h\left(2\csc^2t_-+\frac12\csc^2(t_-/2)-\eta\sin t_-\right)>0.
\end{equation}
\end{proposition}

\begin{proof}
For $t\in(0,\pi)$ let
\[
V_\pm(t):=(1-\cos t)^2(1+\cos t)e^{\pm\eta\sin t}
=2\sin^2t\sin^2(t/2)e^{\pm\eta\sin t}.
\]
Their logarithmic derivatives are
\begin{align}
 g_\pm(t):=\frac{V_\pm'(t)}{V_\pm(t)}
 =2\cot t+\cot(t/2)\pm\eta\cos t,
 \label{eq:g}
\end{align}
and
\begin{align}
 g_\pm'(t)=-2\csc^2t-\frac12\csc^2(t/2)\mp\eta\sin t<0.
 \label{eq:gprime}
\end{align}
The strict inequality follows from $\eta=1/2$, since the two negative
terms have absolute sum at least $5/2$. Moreover,
$g_\pm(t)\to+\infty$ at $0$ and $g_\pm(t)\to-\infty$ at $\pi$.
Thus each arc has exactly one maximum; let $t_+$ be the zero of $g_+$.
For every $t\in(0,\pi)$, $V_-(t)<V_+(t)$, and hence
\[
 h=V_-(t_-)<V_+(t_-)\le V_+(t_+)=:h_+.
\]

Any continuous path from $K_A$ to $K_B$ projects to a path from $0$
to $\pi$ on the $x$-circle. A continuous lift of this projection must
cross one of the two arcs from $0$ to $\pi$ and therefore cross either
$-t_-$ or $t_+$. Since $L(x,y)\ge V(x)$, its maximum is at least $h$.
The path
\[
 \gamma_*(s):=(-\pi s,0)\pmod{2\pi},\qquad 0\le s\le1,
\]
joins $(0,0)\in K_A$ to $(-\pi,0)=(\pi,0)\in K_B$ and satisfies
\[
 \max_{s\in[0,1]}L(\gamma_*(s))=V_-(t_-)=h,\qquad
 \gamma_*([0,1])\cap\{L=h\}=\{z\}.
\]
Any path with maximum at most $h$ cannot cross $t_+$; when
it crosses $x=-t_-$ it must satisfy
$a(-t_-)(1-\cos y)=0$. Since $a(-t_-)>0$, it passes through $z$.

At $y=0$, $\partial_yL(x,0)=0$, and \eqref{eq:troot} gives
$\nabla L(z)=(V'(-t_-),0)=(0,0)$.
The mixed Hessian entry at $y=0$ is zero and
$\partial_y^2L(x,0)=a(x)$. Equations \eqref{eq:g}--\eqref{eq:gprime}
give $V''(-t_-)=h g_-'(t_-)=-\mu<0$. This proves
\eqref{eq:model-hessian} and nondegeneracy.
Thus every $\gamma\in\mathcal P_h$ passes through $z$, while
$\gamma_*$ meets $\{L=h\}$ only at $z$. Lemma~\ref{lem:gate-criterion}
now gives the unique gate $\{z\}$, and
\eqref{eq:model-hessian} shows that $z$ has Morse index one.
These facts verify all parts of Assumption~\ref{ass:single-gate}.
\end{proof}

\subsection{Transition time}\label{sec:model-cap}

Proposition~\ref{prop:gate} verifies Assumption~\ref{ass:single-gate},
and Proposition~\ref{prop:model-mass} verifies
\eqref{eq:zeros} and \eqref{eq:volume}. Theorem~\ref{thm:main}
therefore applies to this model.

The volume coefficient and the capacity now give a complete instance
with a nonisolated well and a logarithmic correction.

\begin{corollary}[Transition law with a logarithmic prefactor]\label{cor:example}
For the potential \eqref{eq:model}, let
$B_\rho=\{w:\dist(w,(\pi,0))\le\rho\}$ with $\rho>0$ sufficiently
small and fixed. For every nonempty compact $Q\subset U$,
\begin{equation}\label{eq:final-model}
 \E_{(x,y)}\left[\tau_{B_\rho}\right]
 =\sqrt{\frac{\pi\kappa}{\mu}}\,
 \eps^{-1/2}\log(1/\eps)e^{h/\eps}(1+o(1)),
 \qquad (x,y)\in Q,
\end{equation}
with relative error uniform on $Q$. Here $h,\mu,\kappa$ are given in
\eqref{eq:troot}--\eqref{eq:model-hessian}.
\end{corollary}

\begin{proof}
Proposition~\ref{prop:model-mass} gives $a_A=2$, $\lambda_A=1/2$,
$m_A=2$, and $c_A=\sqrt\pi$. By Proposition~\ref{prop:gate},
$|\det H_z|=\mu\kappa$ and $\mu_-=\mu$; since $d=2$,
\eqref{eq:cz} gives $c_z=\sqrt{\mu/\kappa}$. Their ratio in
Theorem~\ref{thm:main} is $\sqrt{\pi\kappa/\mu}$.
\end{proof}

\end{document}